\documentclass[preprint,review,12pt]{elsarticle}

\usepackage{amssymb}
\usepackage{amsmath}
\usepackage{subfigure}
\usepackage{pifont}
\usepackage{url}
\usepackage{multirow}
\usepackage{float}
\usepackage{makeidx}
\usepackage{tabularray}
\makeindex
\usepackage{arydshln}
\usepackage{xcolor}

\journal{Biomedical Signal Processing and Control}
\begin{document}
\printindex
\begin{frontmatter}



\title{Unified Efficient Hierarchical Swin Attention Networks for Low-Dose PET and CT Denoising}

\author[label1]{Yichao Liu} 
\author[label2]{Hengzhi Xue}
\author[label2,label3]{Yueyang Teng\corref{cor1}}
\ead{tengyy@bime.neu.edu.cn}
\author[label4]{Junwen Guo\corref{cor1}}
\ead{junwen.guo@umu.se}
\cortext[cor1]{Corresponding author}

\affiliation[label1]{organization={IWR, Heidelberg University},
	            city={Heidelberg},
	            postcode={69120}, 
	            state={Baden Württemberg},
	            country={Germany}}

\affiliation[label2]{organization={College of Medicine and Biological Information Engineering, Northeastern University},
	            city={Shenyang},
	            postcode={110169}, 
	            state={Liaoning},
	            country={China}}
\affiliation[label3]{organization={Key Laboratory of Intelligent Computing in Medical Image, Ministry of Education},
	            city={Shenyang},
	            postcode={110169}, 
	            state={Liaoning},
	            country={China}}     

\affiliation[label4]{organization={Department of Epidemiology \& Global Health, Umeå University},
	addressline={}, 
		city={Umeå},
		postcode={90187}, 
		country={Sweden}}

\begin{abstract}

Low dose computed tomography (CT) reduces radiation exposure to approximately 70-80\% of the standard diagnostic CT, improving medical operation safety while losing details and introducing noise on the resulting imaging. For positron emission tomography (PET), it decreases the tracer dose to 1\%. Supervised vision models and generative technologies enhance these poor-quality diagnostic imageries, but suffer from trade-offs between denoising performance, optimization robustness, and computational efficiency. We aim to improve Swin transformer based denoising solutions to achieve visible reduction of computational costs while enhancing performance. We leverage efficient spatial attention and a hierarchical structure to decrease model parameters and memory consumption. To this end, we propose Efficient Hierarchical Swin Attention Network (EHSANet), a framework with an improved multi-stage feature encoding-decoding structure integrating efficient global attention modules (EGAM), in addition to a hybrid fusion upsampling module (HFUM) to perform resolution recovery for supervised denoising. Extensive experiments show that our model reduces FLOPs (floating point operations) and GPU memory compared to SwinIR and offers improved trade-offs compared to a few state-of-the-art methods. Meanwhile, our model demonstrates visible improvement over optimization stability with low variance and offers denoising enhancement.

\end{abstract}



\end{frontmatter}


\section{Introduction}
\label{Introduction}
Medical imaging modalities such as computed tomography (CT) and positron emission tomography (PET) are indispensable tools for the diagnosis, staging, and monitoring of a wide range of diseases. However, repeated or high-dose exposure to ionizing radiation poses significant health risks to patients, particularly in routine screening, pediatric imaging, or longitudinal follow-up \cite{brower2021radiation, hirshfeld20182018}. To mitigate these risks, low-dose imaging protocols have been widely adopted \cite{immonen2022use, vonder2021latest}. While effective in reducing radiation exposure, these protocols often lead to images with increased noise, reduced contrast, and compromised diagnostic quality \cite{clement2025ai, zubair2024enabling, caruso2024low}. Consequently, developing robust denoising techniques that can preserve anatomical and functional fidelity while suppressing noise has become a critical area of research. Recent advances in deep learning and data-driven reconstruction methods offer promising solutions to this challenge, enabling high-quality image recovery from low-dose acquisitions across CT and PET modalities \cite{xue2025noise, lei2019whole, chen2017low, wang2023ctformer}. 

\begin{figure}[!ht]
\centering
		\includegraphics[width=0.6\textwidth]{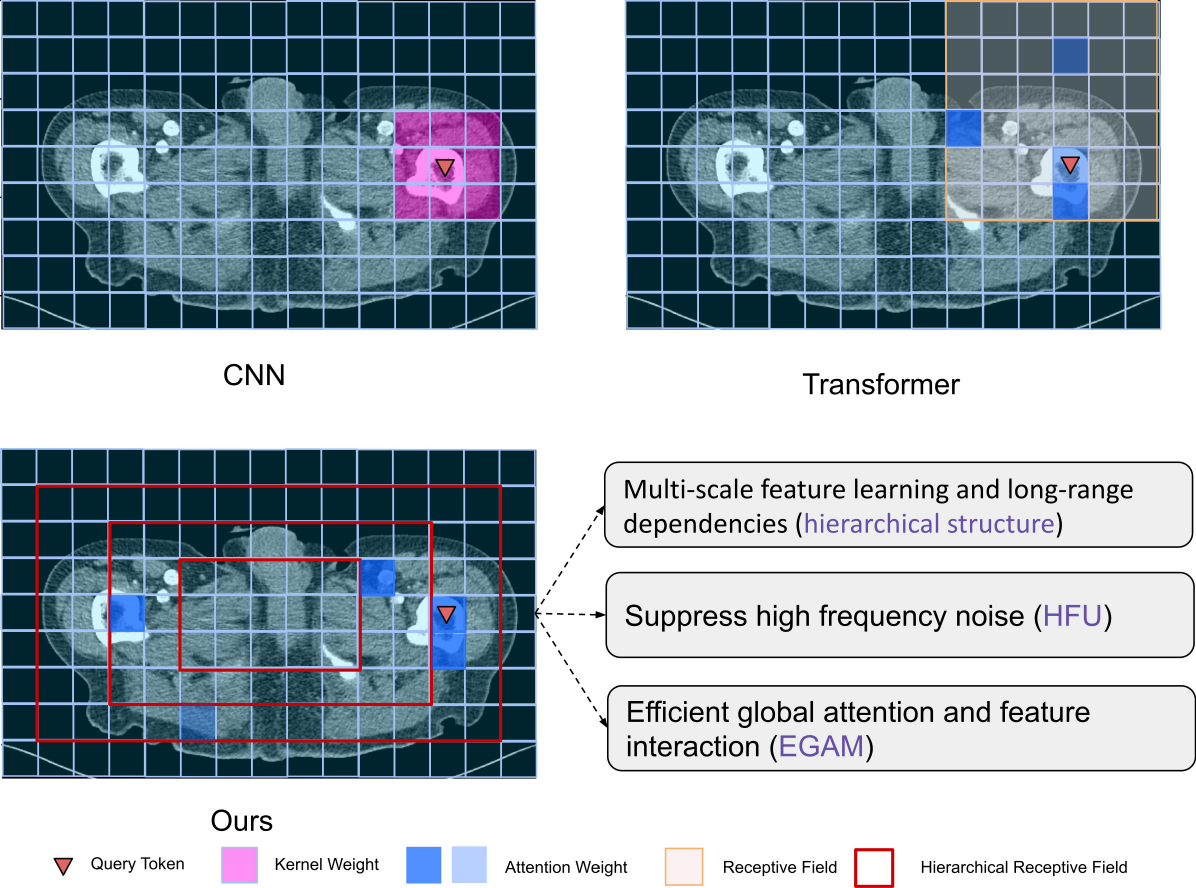}
    \caption{Motivation illustration. Local receptive field of CNNs and restricted receptive long-range dependencies of Swin Transformer limit their ability to capture global and multi-scale features. Our proposed method can capture  semantic information, hierarchical structure information and reduce computational cost.}
    \label{motivation}
\end{figure}

Especially, with deep learning has dominated image denoising research due to their spatial locality and efficiency \cite{zheng2021deep, lefkimmiatis2017non, lefkimmiatis2018universal}, lots of works have been applied in low-dose CT (LDCT) and LDPET denoising \cite{heinrich2018residual, chen2017low, li2020investigation}. However, the inherently local receptive field of convolution limits their capacity to model long-range dependencies \cite{wang2018non}. This limitation has motivated the exploration of architectures capable of capturing global contextual relationships, as illustrated in Fig. \ref{motivation}. Recently, self-attention–based models, particularly Transformer architectures, have demonstrated superior performance in medical image denoising tasks \cite{luthra2021eformer, wang2023ctformer}, owing to their ability to model global interactions explicitly. Nevertheless, both CNNs and Transformers exhibit optimization instability under certain training regimes \cite{xiao2021early}. To mitigate these issues, hybrid architectures have been proposed, leveraging the strong inductive bias of convolutional layers for early-stage feature extraction and self-attention mechanisms for high-level global modeling, resulting in improved performance and enhanced training stability \cite{liang2021swinir}. Despite these advances, hybrid models typically incur substantial computational overhead and prolonged training time \cite{sutariya2025architectural,nagaraju2024low,lu2022transformer,li2024grformer}. To address these efficiency constraints, hierarchical modeling strategies have been widely adopted to optimize CNN computations \cite{ronneberger2015u, qin2020u2, jin2023novel, zhang2022real}. By progressively downsampling feature maps, hierarchical architectures enable multi-scale representation learning, integrating fine-grained local textures at high resolutions with broader structural context at coarser scales. This design substantially reduces overall FLOPs while preserving strong restoration capability.

\begin{figure}[!ht]
\centering
		\includegraphics[width=0.5\textwidth]{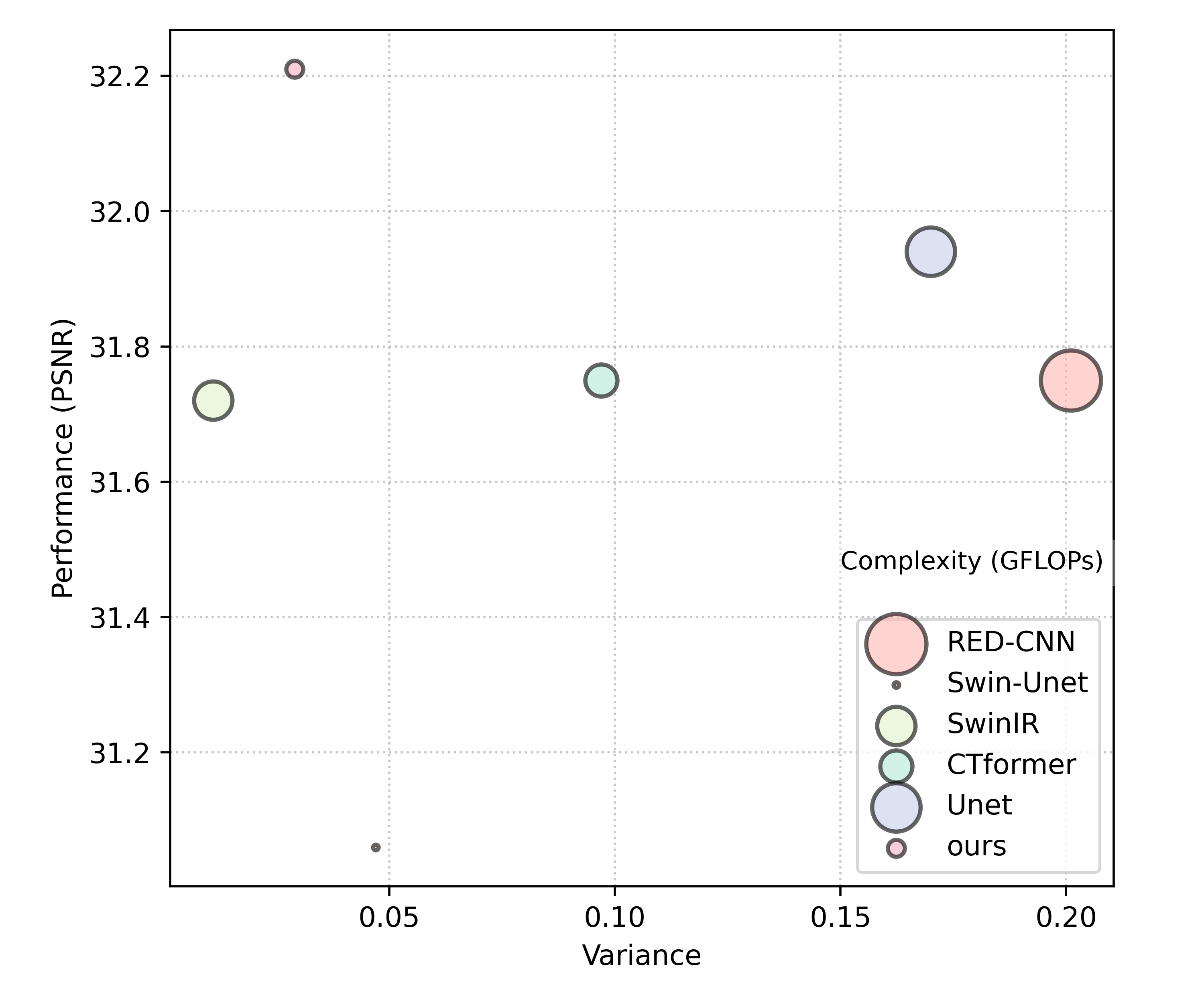}
    \caption{Comparison of PSNR, FLOPs and training performance variance between ours and others on AAPM LDCT dataset, where the larger circle represents more FLOPs.}
    \label{figure1}
\end{figure}

Various of studies show that hierarchical structure utilizing multi-scale features can achieve better performance on image denoising \cite{prakash2021interpretable, wang2022uformer, li2023h2tf}. In this study, we propose to leverage the advantage of hierarchical structure and design an Efficient Hierarchical Swin Attention Network (EHSANet) for LDCT/PET image denoising based on SwinIR. The goal of this architecture is to be able to capture both local and long-range dependencies, as well as the hierarchical structure of medical images with limited training resources. To further enhance generalization and robustness, we introduce a novel patch expanding block, which focuses on learning from low-frequency textures, helping the model avoid overfitting to high-frequency noise. This is inspired by Deep Image Prior (DIP) \cite{ulyanov2018deep, liu2023devil}, which offers a complementary architecture level design that upsampling can significantly affect denoising performance. In addition, we improve our previous global attention mechanism by drastically reducing its parameter count, making it significantly more efficient. Our experiments demonstrate that EHSANet outperforms mainstream denoising methods, achieving superior denoising performance with fewer parameters and comparable GPU memory consumption, making it highly suitable for practical applications, as shown in Fig. \ref{figure1}.

The main contributions of the papers are as follows:
\begin{itemize}
	\item We propose a novel LDCT/PET denoising model, EHSANet, which is enhanced by capturing long and short range dependency, structure feature extraction and texture preservation. This hybrid design leads to more stable optimization, lower computation cost and consistently improved denoising results.
	\item We design a Hybrid Fusion Upsampling (HFU) module, which encourages the model to focus on low-frequency image structures, helping to suppress overfitting to high-frequency noise.
	\item We introduce an efficient attention module, EGA, which comprises two variants: the Efficient Global Attention Module-Sequential (EGAM-S) and Efficient Global Attention Module-Parallel (EGAM-P). They are embedded in feature extraction and HQ reconstruction, as well as encoder-decoder block. Together, they enhance information fusion between the encoder and decoder, suppress noise, and achieve higher Peak Signal-to-Noise Ratio (PSNR) with significantly fewer parameters.
\end{itemize}
\section{Related work}
\subsection{CNN and transformer-based medical image denoising}
With the rapid development of image restoration, super resolution, image reconstruction and denoising, various technologies have been adapted to LDCT/PET denoising. RED-CNN \cite{chen2017low} utilizes a convolutional autoencoder architecture with residual learning to effectively suppress noise in LDCT images while preserving structural details. Cycle-consistent generative adversarial network (CycleGAN), along with its variants Identity GAN and GAN-CIRCLE, were investigated by Li \textit{et al.} \cite{li2020investigation} for unpaired LDCT denoising. These approaches effectively learn image translation from the low-dose to the full-dose domain without the need for aligned training data. Transformer-based architectures, originally developed for natural language processing \cite{vaswani2017attention}, have recently shown remarkable success in various computer vision tasks by leveraging self-attention mechanisms to model global context \cite{dosovitskiy2020image}. CTformer \cite{wang2023ctformer} is a fully convolution-free architecture for LDCT denoising that leverages a Token2Token dilated vision transformer to effectively model both local textures and global anatomical structures through hierarchical self-attention. While Vision Transformers (ViTs) demonstrate strong global modeling capabilities through full self-attention, their fixed patch structure and high computational cost limit their effectiveness in dense prediction tasks such as CT denoising. To overcome these issues, Swin Transformer \cite{liu2021swin} introduces a hierarchical, shifted-window self-attention mechanism that enables both scalability and improved local-context modeling. STEDNet \cite{zhu2023stednet} incorporates a noise extraction subnetwork and hierarchical multi-scale attention to effectively suppress noise and artifacts while preserving structural detail and contrast. 

\subsection{Hybrid model for medical image denoising}
In a recent study, it shows that replacing the standard patchify stem with early convolutions significantly improves the training stability and optimizability of Transformers, making them less sensitive to hyperparameter choices \cite{xiao2021early}. Inspired by this, SwinIR \cite{liang2021swinir} adapts the Swin Transformer for image restoration by integrating a shallow convolutional feature extraction module and residual connections, making it highly effective for denoising applications. Wang \textit{et al.} \cite{wang2023low} present a comprehensive evaluation of five state of the art AI models for restoring low-count whole-body PET$/$MRI images across a wide dose reduction spectrum, demonstrating that SwinIR consistently achieves superior quantitative and qualitative performance, particularly at practically significant ultra-low-dose levels. Besides, a more recent study also shows that SwinIR is comparable with state of art models on standard RGB dataset \cite{he2024training}. Although SwinIR shows superior performance and enhanced training stability, based on the result from the study \cite{wang2023low}, the model consumes huge training memory and higher floating point operations (FLOPs).


\section{Methods}
\label{Methods}
\subsection{Network architecture}
\begin{figure}[H]
	\centering
	\subfigure[EHSANet]{
		\includegraphics[width=0.85\textwidth]{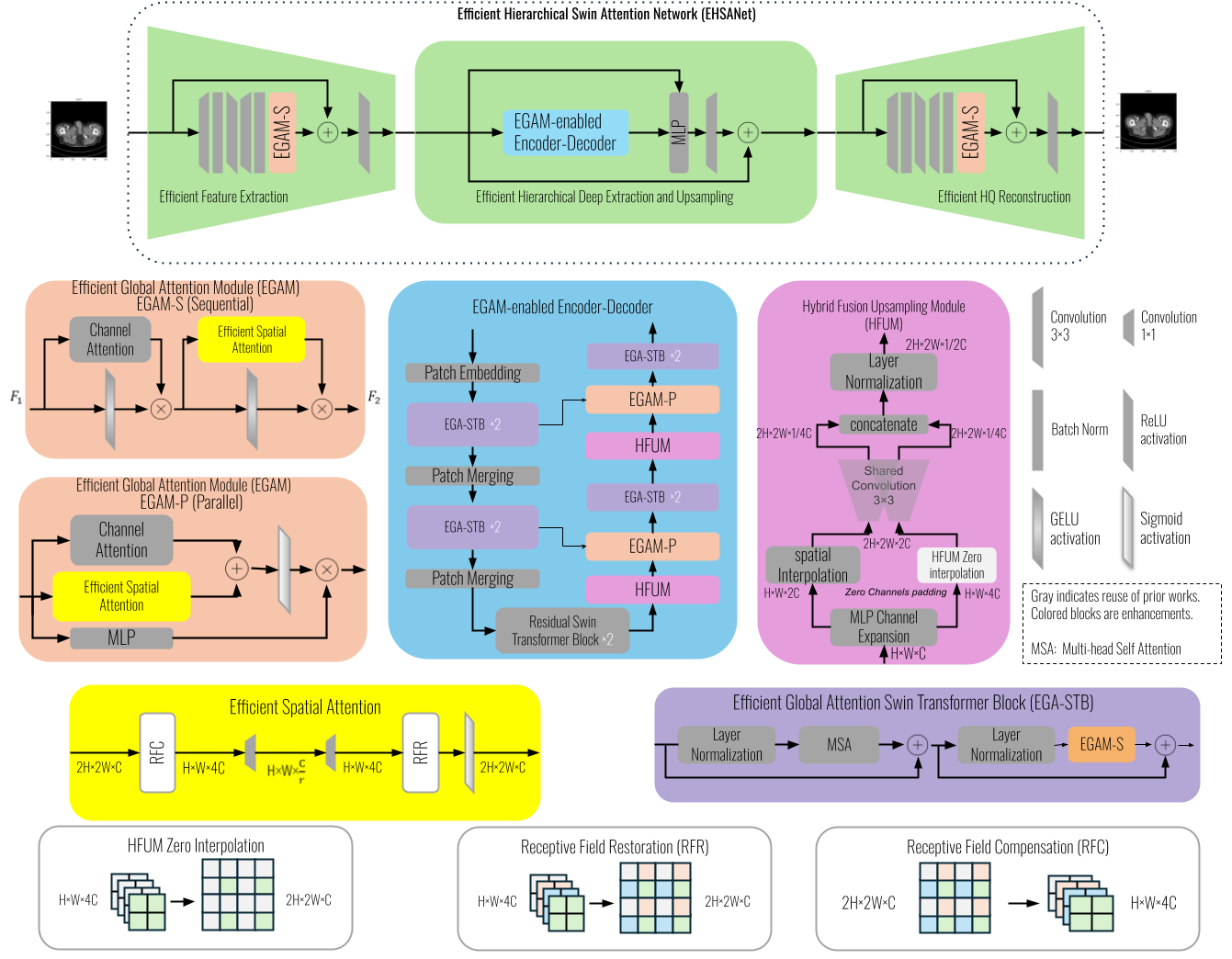}
	}
	  
	\caption{The structure of EHSANet. It consists of feature extraction and reconstruction blocks, along with a central residual encoder-decoder block. The proposed Efficient Sequential Global Attention (E(S)GA) and Efficient Parallel Global Attention (E(P)GA) modules are embedded in those blocks as well as Swin Transformer block. The encoder-decoder block is designed to learn hierarchical representations from LDCT images. The Hybrid Fusion Upsampling (HFU) module is embedded in the decoder side to keep the structure information. We adopt nearest interpolation as spatial interpolation. In the zero interpolation module, white boxes represent zero. We expand feature size by interleaving zeros between columns and rows.}\label{HSANet}
\end{figure}

The architecture of EHSANet (illustrated in Fig. \ref{HSANet}) consists of three main components: efficient feature extraction, efficient hierarchical deep extraction and upsampling and efficient high-quality (HQ) reconstruction modules. The efficient feature extraction and high-quality (HQ) reconstruction modules are used to introduce inductive bias and stablize the optimization \cite{xiao2021early}. The representation is enhanced by embedding our efficient attention module, EGAM-S. The efficient hierarchical deep extraction and upsampling module is responsible for capturing long-range dependencies. Inspired by Swin-Unet \cite{liu2021swin}, an Encoder-Decoder structure is introduced to effectively extract multi-scale structural features and reduce the computation cost. To enhance texture preservation and edge sharpness, we have designed a residual multi-layer perceptron (MLP) structure, which helps retain high-frequency details while enabling effective feature fusion. A convolutional layer with residual structure is adopted at the end of the efficient hierarchical deep extraction and upsampling module to introduce inductive bias to transformer based network and provide a better connection to local dependency and long range dependency. The overall architectural design aims to mitigate the high computational costs and memory overhead associated with the stacked transformer bottlenecks found in SwinIR. We propose two types of Efficient Global Attention modules, EGAM-S and EGAM-P. These modules address the over-parameterization of our previous method, Global Attention Mechanism (GAM)\cite{liu2021global}. The proposed attention modules are used for richer feature interaction and to suppress redundant or noisy information. Details of these modules are provided in Section \ref{att}. Finally, we have proposed a hybrid fusion upsampling module, designed to learn low-frequency structures such as background tissue, organs, and bones, instead of focusing on high-frequency noise. Further details are provided in Section \ref{upsampling}.

\subsection{Attention module}
\label{att}
\subsubsection{EGAM-S module}
Our previous GAM model \cite{liu2021global} demonstrated significant performance improvements over the CBAM \cite{woo2018cbam} and BAM \cite{park2018bam} modules. However, the number of parameters increases significantly as the model size grows. To address this, we propose an efficient spatial attention module and replace the original $7 \times 7$ convolution kernel with a more lightweight $1 \times 1$ convolution, effectively reducing the number of parameters. To compensate for the reduced receptive field and to capture broader spatial patterns and contextual features, we propose a receptive field compensation (RFC) operation, which is an (inverse) pixel shuffle operation in \cite{shi2016real}, and a corresponding receptive field restoration (RFR) operation. Fig. \ref{HSANet} shows RFC, RFR, spatial efficient attention module as well as the EGAM-S module, and Eq. (\ref{ESGA_c}) and Eq. (\ref{eq_ESGA}) show details. Specifically, let the input feature be $F_{1} \in \mathbb{R}^{N\times H\times W\times C}$. For channel attention, we employ an encoder-decoder MLP structure to enhance channel-wise information interaction, which is formulated as follows:
\begin{equation}
	F_{channel}=F_{1} \circledast sigmoid(L_{2}(L_{1}(F_{1})))
\label{ESGA_c}
\end{equation}
where, $L_{1}$ stands for a linear layer with weight, $W_{1} \in \mathbb{R}^{C\times C/r}$, while $L_{2}$ stands for a linear layer with weight, $W_{2} \in \mathbb{R}^{C/r\times C}$. $r$ is a compression ratio. $\circledast$ denotes element-wise multiplication.

For spatial attention, an encoder-decoder $1 \times 1$ convolution structure is used to improve spatial-wise information interaction (Fig. \ref{HSANet} RFR and RFC module), which can be represented as below:
\begin{equation}
	F_{2}=F_{channel} \circledast sigmoid(\pi^{-1}(\mathcal{PS}(Conv_{2}(Conv_{1}(\pi(\mathcal{PS}^{-1}(F_{channel})))))))
\label{eq_ESGA}
\end{equation}
where $\mathcal{PS}$ is periodic shuffling operator, which is introduced for the modified version of the GAM method. More detailed information can be found in \cite{shi2016real}. $\mathcal{PS}^{-1}$ is the inverse of periodic shuffling operator. $\pi$ and $\pi^{-1}$ are permutation and inverse permutation operations. $Conv_{1}$ and $Conv_{2}$ are convolution kernels introduced for the modified GAM as well, $k_{1} \in \mathbb{R}^{4C\times C/r\times 1\times 1}$, $k_{2} \in \mathbb{R}^{C/r\times4C\times 1\times 1}$, separately. Note that the integration of periodic shuffling operator and $1 \times 1$ convolution structure minimizes both parameters and computation cost.

To enable the Swin Transformer block to capture both long-range and short-range dependencies, we replace the MLP component with EGAM-S module, incorporating GELU activation \cite{hendrycks2016gaussian}. \label{app1} Fig. \ref{HSANet} EGA-STB module shows the efficient global attention swin transformer block (EGA-STB).

\subsubsection{EGAM-P module}
We adapt the EGAM-S module in a parallel manner at the skip connections in the decoder to enhance channel and spatial interactions while suppressing noise from the encoder, as shown in Fig. \ref{HSANet}. Assume the input feature is $F_{cat} \in \mathbb{R}^{N \times H \times W \times 4C}$, the output feature is calculated as below:

\begin{multline}
	F_{out}=sigmoid(0.5*L_{2}(L_{1}(F_{cat}))+\\
	0.5*\pi^{-1}(\mathcal{PS}(Conv_{2}(Conv_{1}(\pi(\mathcal{PS}^{-1}(F_{cat}))))))) \circledast L(F_{cat})
\end{multline}
where $L$ is MLP layer with weight $W \in \mathbb{R}^{2C \times C}$

\subsection{Hybrid Fusion Upsampling module}
\label{upsampling}
In Swin-Unet \cite{cao2022swin}, the patch expanding module uses a MLP to expand the channel dimension from $C$ to $4C$, followed by a rearrangement operation, similar to periodic shuffling, to increase the spatial resolution from $H \times W$ to $2H \times 2W$. However, as noted in \cite{liu2023devil}, such learned upsampling strategies often tend to overfit to noise. To address this issue, we propose HFU module to force network to learn from low-frequency upsampling interpolation strategies. As illustrated in Fig. \ref{HSANet}, our HFU module first uses a MLP for channel expansion. The resulting feature map is then split into two branches: one undergoes nearest neighbor interpolation, while the other applies zero-padding interpolation operation. A shared convolutional layer is subsequently applied to both branches to ensure the network learns interpolation-invariant features, enhancing robustness and generalization.

\subsection{Loss function}
We have introduced Mean Absolute Error (MAE) loss function. We choose MAE as loss function, as it can capture high-frequency details \cite{huang2022deep}. Byeongjoon \textit{et al.} have compared MAE and MSE loss for CT image denoising\cite{kim2019performance}. It shows that both MAE and MSE are commonly used, however, both of them fail to capture semantic details. In our study, MAE works better than MSE loss. Thus, the final loss function is formulated as:

\begin{equation}
	L_{denoise}=\frac{1}{N}\sum_{i=1}^{N}\frac{1}{HW}|X_{LD}^{i}-X_{FD}^{i}|_{1}
\end{equation}

\section{Experiments}
\subsection{Dataset and evaluation}
\subsubsection{Mayo abdomen data}
The AAPM Low-Dose CT Grand Challenge released a widely used public dataset containing paired LDCT and FDCT images, covering the head, chest, and abdomen. For this study, we selected the abdominal data, where each CT volume has a voxel size of $0.5859 \times 0.5859 \times 3.0 mm^{3}$ and an axial resolution of $512 \times 512$. A total of 27 patients were used for training and 9 patients for testing. During preprocessing, the provided DICOM images were first converted to Hounsfield Units (HU), then normalized to the range [0, 1] using a fixed HU window of [-1024, 3072]. 
\subsubsection{Mayo chest data}
The chest cohort consisted of 50 patients scanned using a Siemens SOMATOM Definition AS+ scanner. The routine-dose data were subsequently processed to simulate a 10\% dose level, with images reconstructed at a slice thickness of 1.5 mm. From this cohort, 27 patients were randomly selected for training and 9 patients were reserved for testing.

\subsubsection{UDPET data}
The Ultra-Low-Dose PET Imaging (UDPET) Challenge provides a rich and comprehensive dataset designed to advance research in PET image reconstruction under reduced dose settings. PET data used in our work were obtained from the University of Bern, Dept. of Nuclear Medicine and School of Medicine \cite{xue2022cross}. The dataset comprise 1,447 whole-body 18F-FDG PET scans, acquired across total-body PET systems, Siemens Biograph Vision Quadra: 387 subjects, United Imaging uEXPLORER: 1,060 subjects. From each PET scan, corresponding low-dose images are generated at multiple Dose Reduction Factors (DRFs), specifically at DRFs of 4, 10, 20, 50, and 100, in addition to the original full-dose reference image. Each PET image has an isotropic voxel spacing of 1.65 mm in all three dimensions, with an axial matrix size of 440 $\times$ 440. We randomly pick DRF=100 of 27 patients for training, 9 for testing. We convert the provided DICOM data to standardized uptake values (SUVs) and normalize to the range [0, 1] using min–max scaling. The resulting images were used as input to the network. 

For quantitative evaluation, evaluated model performance using three standard metrics following the implementation of Chen \textit{et al.} \cite{chen2017low}: peak signal-to-noise ratio (PSNR), structural similarity index (SSIM), and root mean squared error (RMSE). For the LDCT dataset, we use the standard deviation as a measure of variability to assess the stability of the models.

\subsection{Implementation details}
We have implemented our model with pytorch 2.7 library and ran the models on two NVIDIA TITAN Xp 12G GPUs. We use Stochastic Gradient Descent (SGD) optimizer with weight decay $1e-4$. The learning rate decays as follows, follows the implementation of Swin-Unet \cite{cao2022swin}.
\begin{equation}
	lr=lr_{base}\cdot (1-\frac{n_{iter}}{M_{total}})^{\gamma}
\end{equation}
where $lr_{base}=0.01$, $n_{iter}$ is the current iterations, $M_{total}$ is the total iterations, $\gamma=0.9$ is decay factor. We have chosen 3000 epochs to train the model in patches. In training, we set mini-batch to 16, patch size to $64 \times 64$. During validation and test, we set batch size to 1, and patch size to $512 \times 512$ and $440 \times 440$, respectively.

\section{Results}
\subsection{Ablation study}
We conduct extensive experiments to evaluate the effectiveness of the proposed method. Specifically, we perform ablation studies on the attention modules and the HIC patch expanding module, with the results presented in Table \ref{att_ablation}. Our results show that ESGA module is the most important module as the performance drops a lot without it. 

\begin{table}[!ht]
\centering
\begin{tabular}{lccccc}
\cline{1-6}
ESGA & \ding{55} & \ding{51} &\ding{51}&\ding{55}&\ding{51}\\ 
EPGA & \ding{55} & \ding{51} &\ding{55}&\ding{51}&\ding{51}\\
HIC & \ding{55} & \ding{55} &\ding{51}&\ding{51}&\ding{51}\\
\cline{1-6}
PSNR & 31.9&32.16 &32.22 & 32.13& 32.29 \\
\cline{1-6}
\end{tabular}
\caption{Ablation study for EHSANet in LDCT dataset. \ding{55} and \ding{51} represent whether EHSANet include those modules. }\label{att_ablation}
\end{table}


\subsection{Model comparison}
We have compared our model with five different methods on LDCT dataset. The compared methods include RED-CNN \cite{chen2017low}, CTformer \cite{wang2023ctformer}, UNet \cite{ronneberger2015u},  Swin-Unet \cite{cao2022swin} and SwinIR \cite{liang2021swinir}. 
\subsubsection{Quantitative evaluation results on LDCT abdomen dataset}
\begin{figure*}[htbp]
	\centering
	\subfigure[]{
		\includegraphics[width=0.3\textwidth,trim=0 0 0 0, clip]{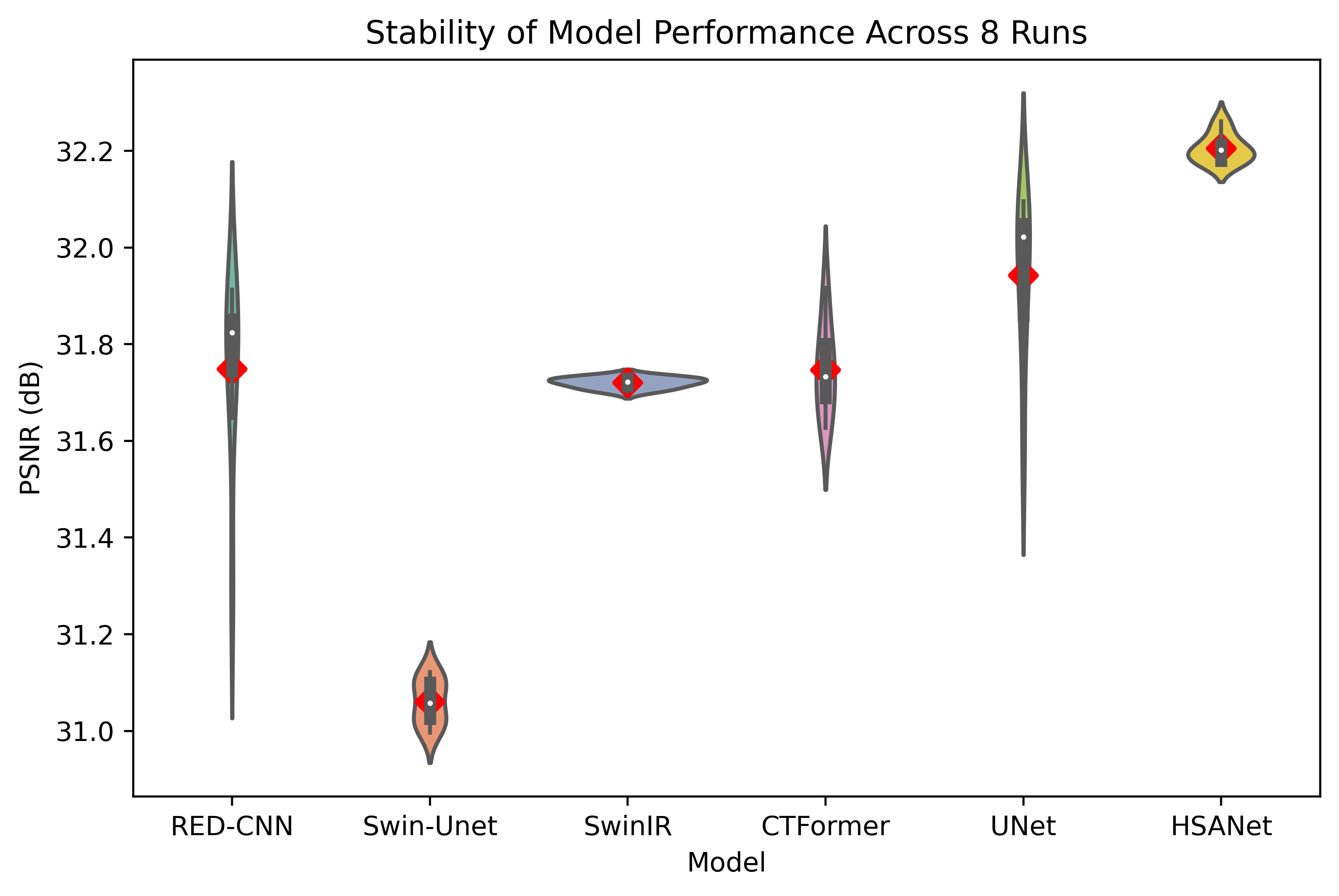}
	}
	\subfigure[]{
		\includegraphics[width=0.3\textwidth,trim=0 0 0 0, clip]{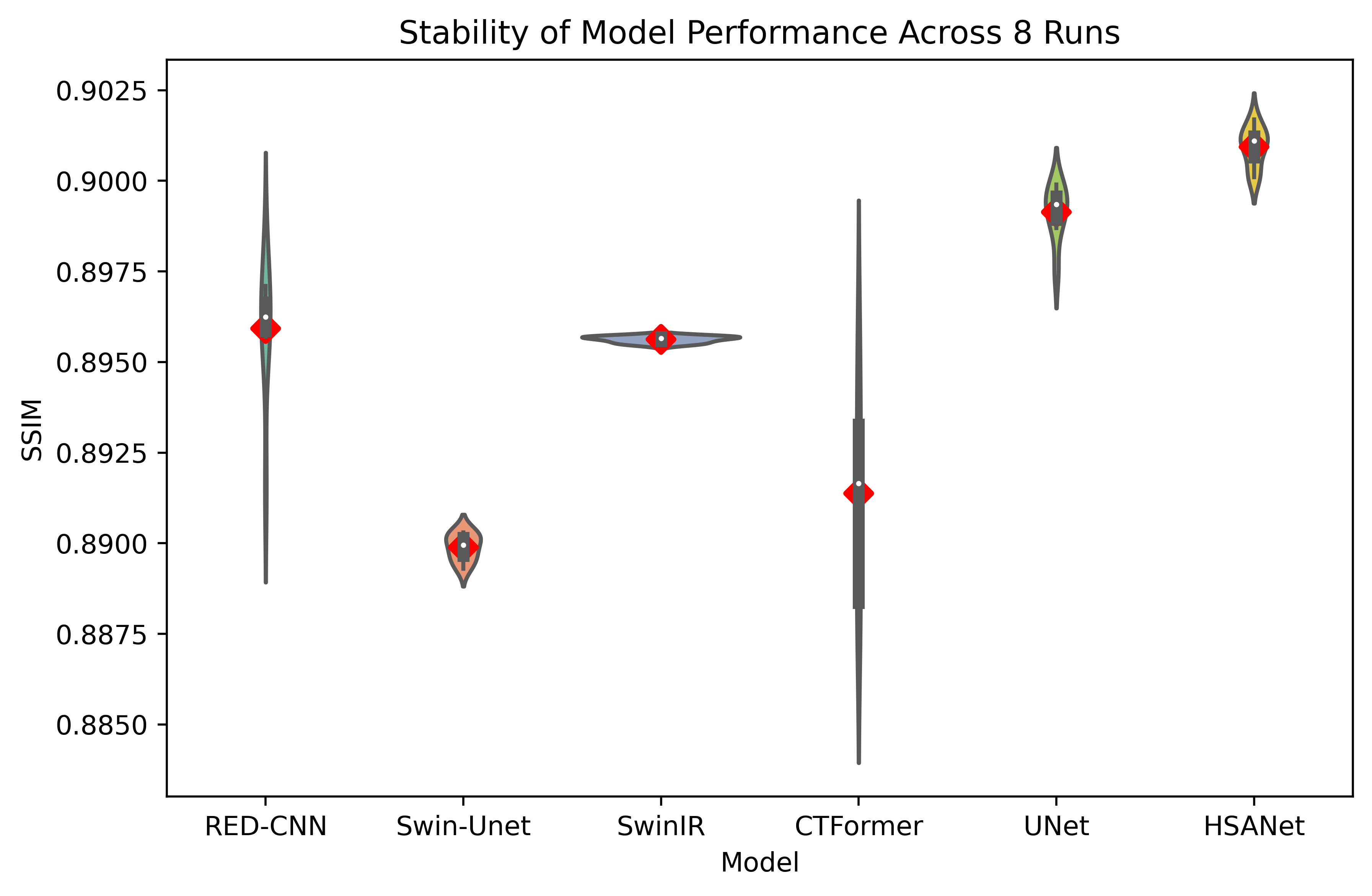}
	}
	\subfigure[]{
		\includegraphics[width=0.3\textwidth,trim=0 0 0 0, clip]{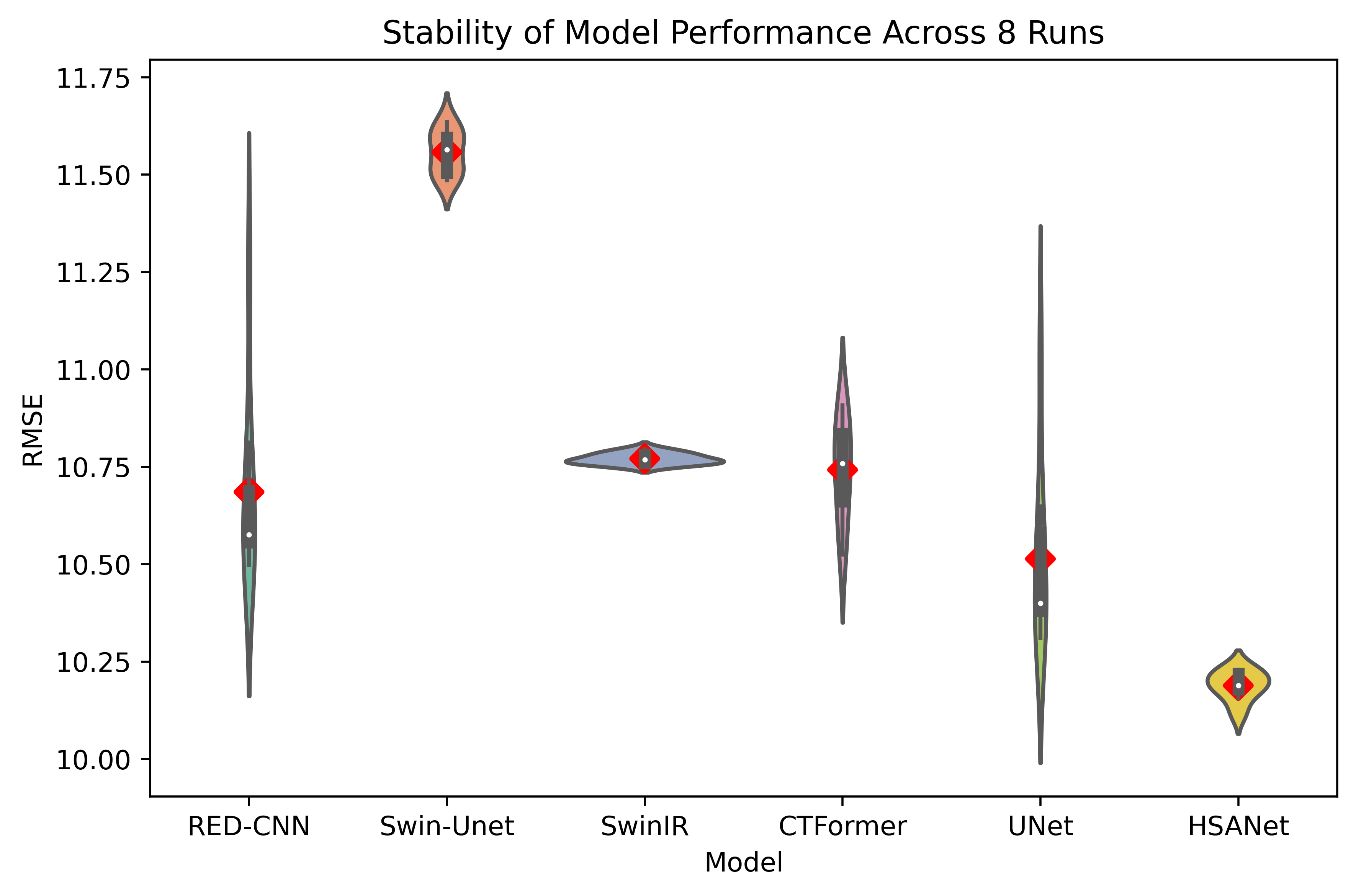}
	}
	\caption{Quantitative (a)PSNR, (b)SSIM and (c)RMSE of different models on 8 different runs. Red points are average. Width of violin plot represent the density of data at each value. Quartiles are shown as thick lines inside the violin plot}
	\label{LDCT_violin}
\end{figure*}

We have adopted the same data preparation procedure as the RED-CNN method to ensure a fair comparison. Specifically, pixel intensities were windowed to [-160, 240] and normalized to [0, 1] before evaluation. The average results from 8 independent runs across the 9 test patients are summarized in Table \ref{comparison}. Our method achieves the best performance in terms of all PSNR, SSIM and RMSE among all compared approaches. Moreover, the variance across runs is minimal, much lower than that of RED-CNN, CTformer, and UNet, and comparable to Swin-Unet for PSNR. In terms of RMSE, our method exhibits even less variability than Swin-Unet. Fig. \ref{LDCT_violin} presents violin plots corresponding to the results in Table \ref{comparison}. The plots reveal that the purely CNN-based approach exhibits the largest variance across runs. In contrast, the transformer-only method (CTformer) shows reduced variance, though still greater than the hybrid CNN-transformer approaches. Our method demonstrates the second-highest stability in terms of PSNR and RMSE, being slightly less stable than SwinIR.

\begin{figure*}[htbp]
	\centering
	\subfigure[]{
		\includegraphics[width=0.22\textwidth,trim=50 10 80 10, clip]{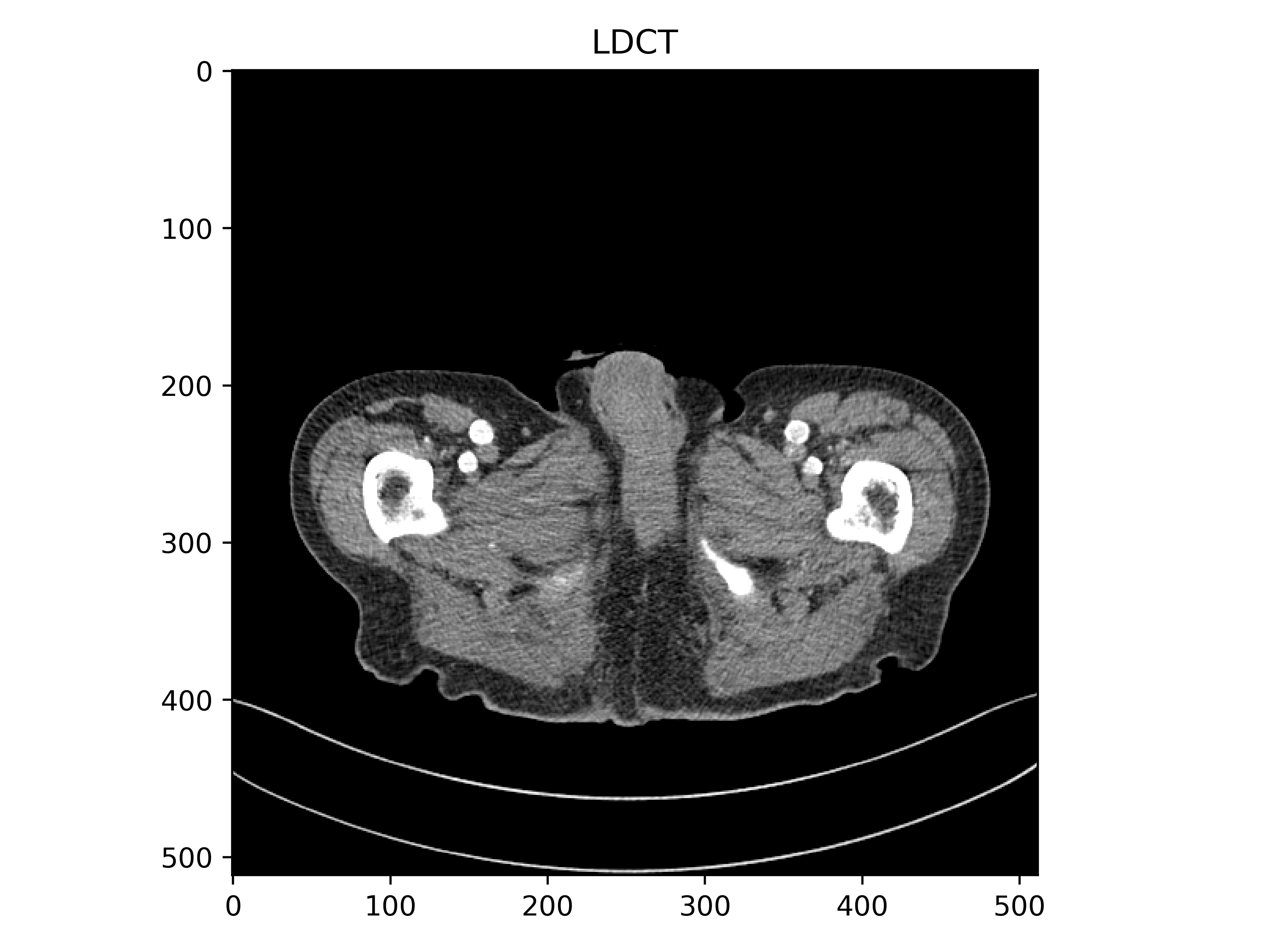}
	}
	\subfigure[]{
		\includegraphics[width=0.22\textwidth,trim=50 10 80 10, clip]{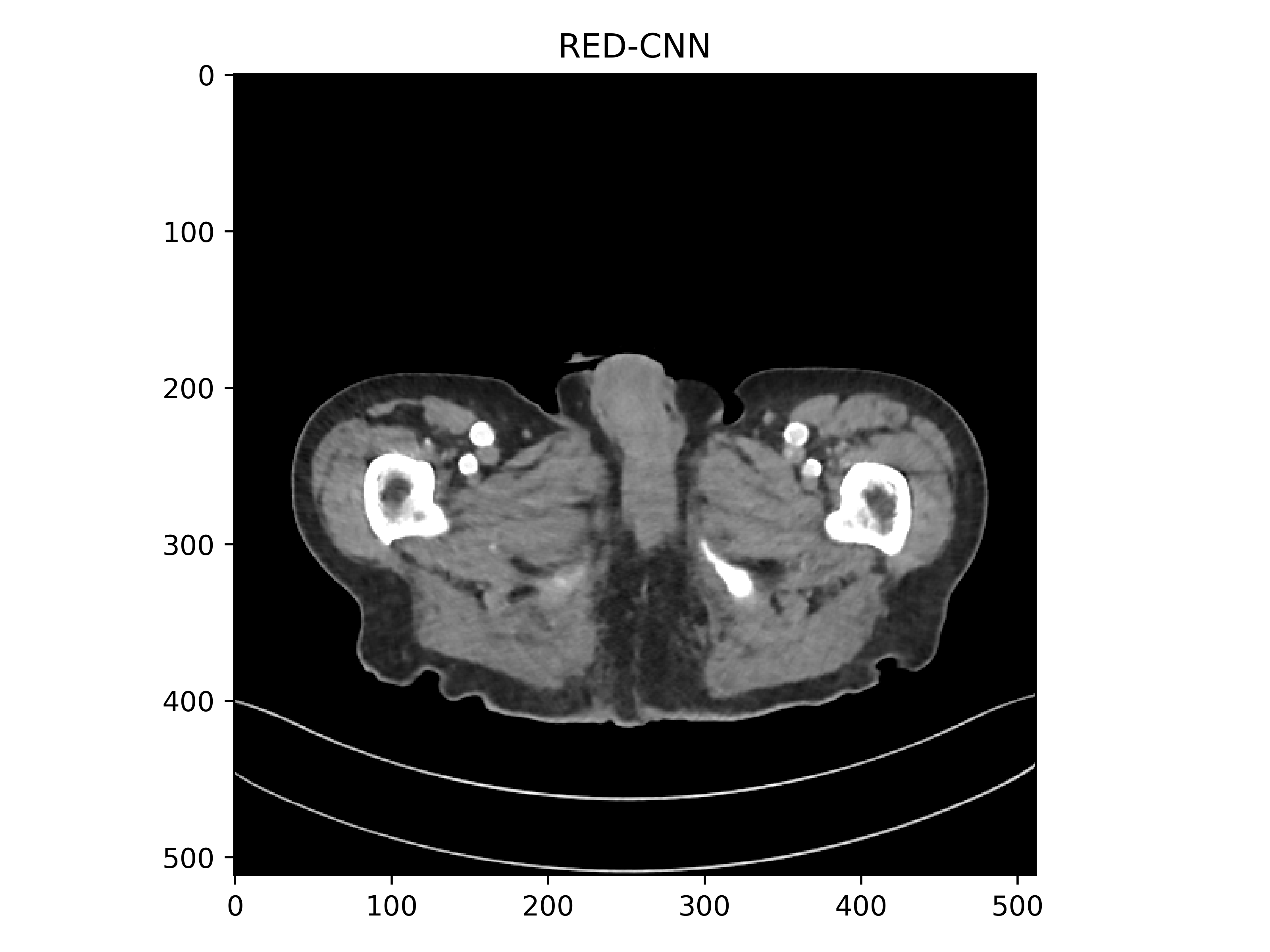}
	}
	\subfigure[]{
		\includegraphics[width=0.22\textwidth,trim=50 10 80 10, clip]{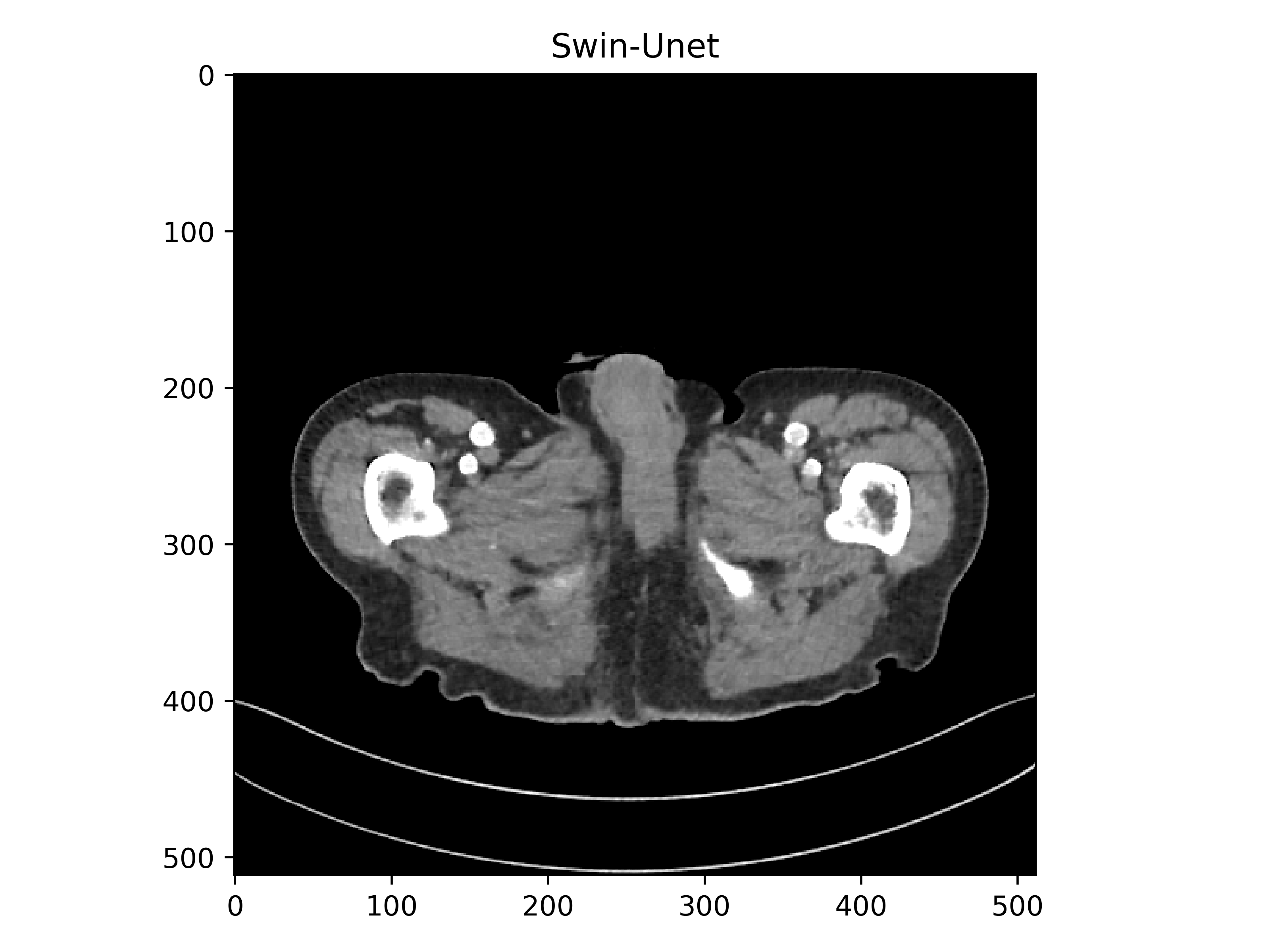}
	}
	\subfigure[]{
		\includegraphics[width=0.22\textwidth,trim=50 10 80 10, clip]{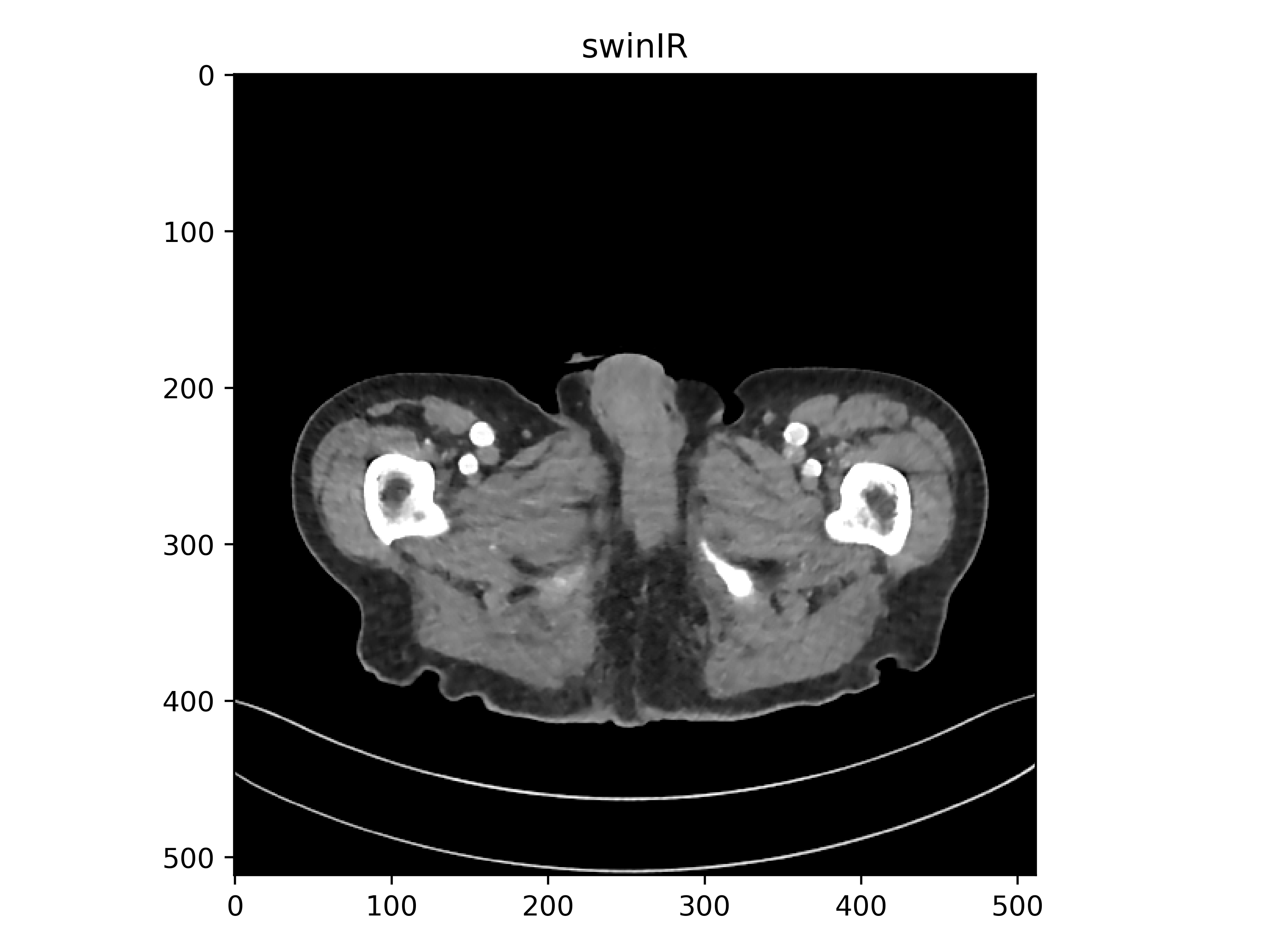}
	}
	\quad
	\subfigure[]{
		\includegraphics[width=0.22\textwidth,trim=50 10 80 10, clip]{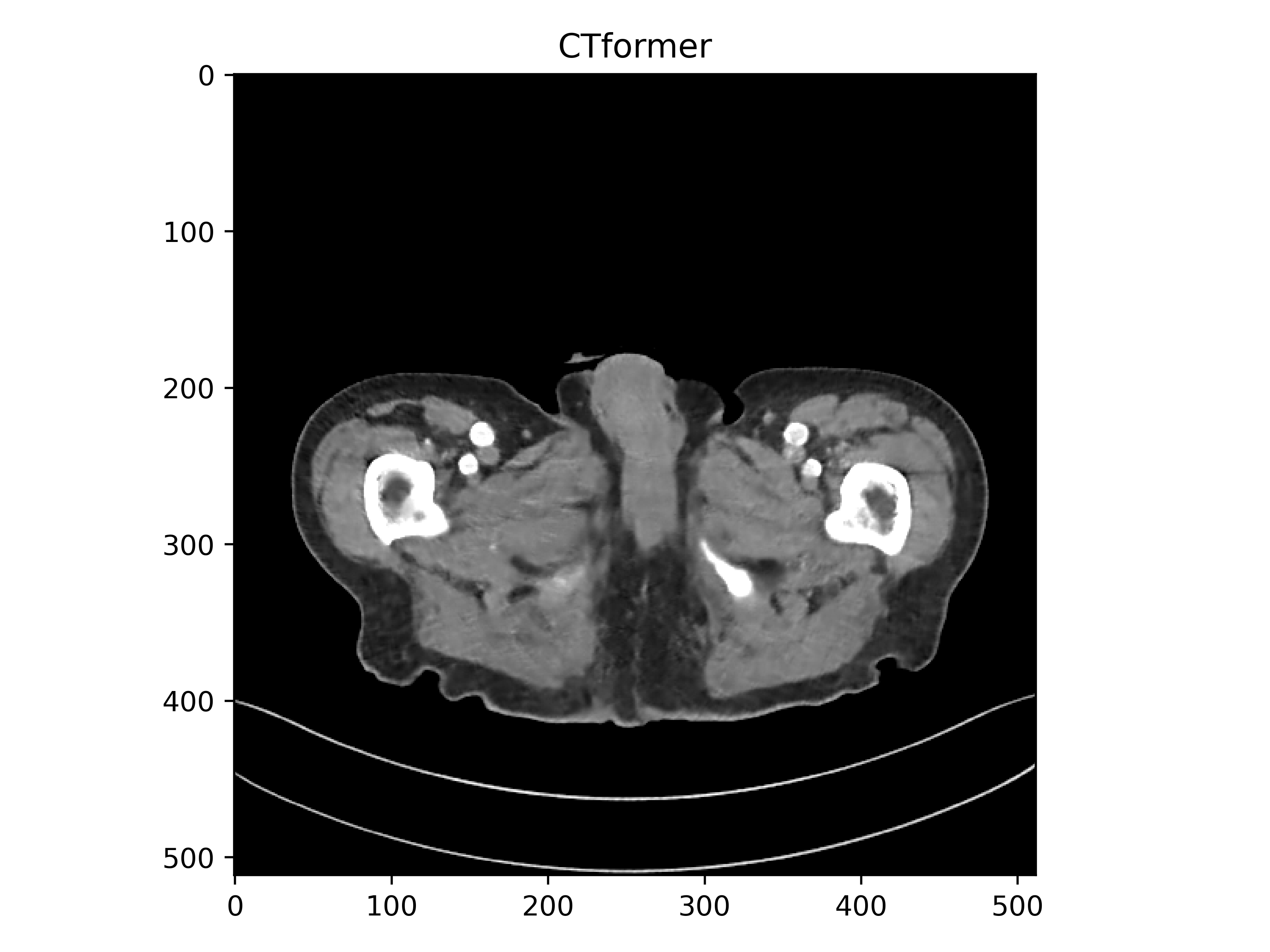}
	}
	\subfigure[]{
		\includegraphics[width=0.22\textwidth,trim=50 10 80 10, clip]{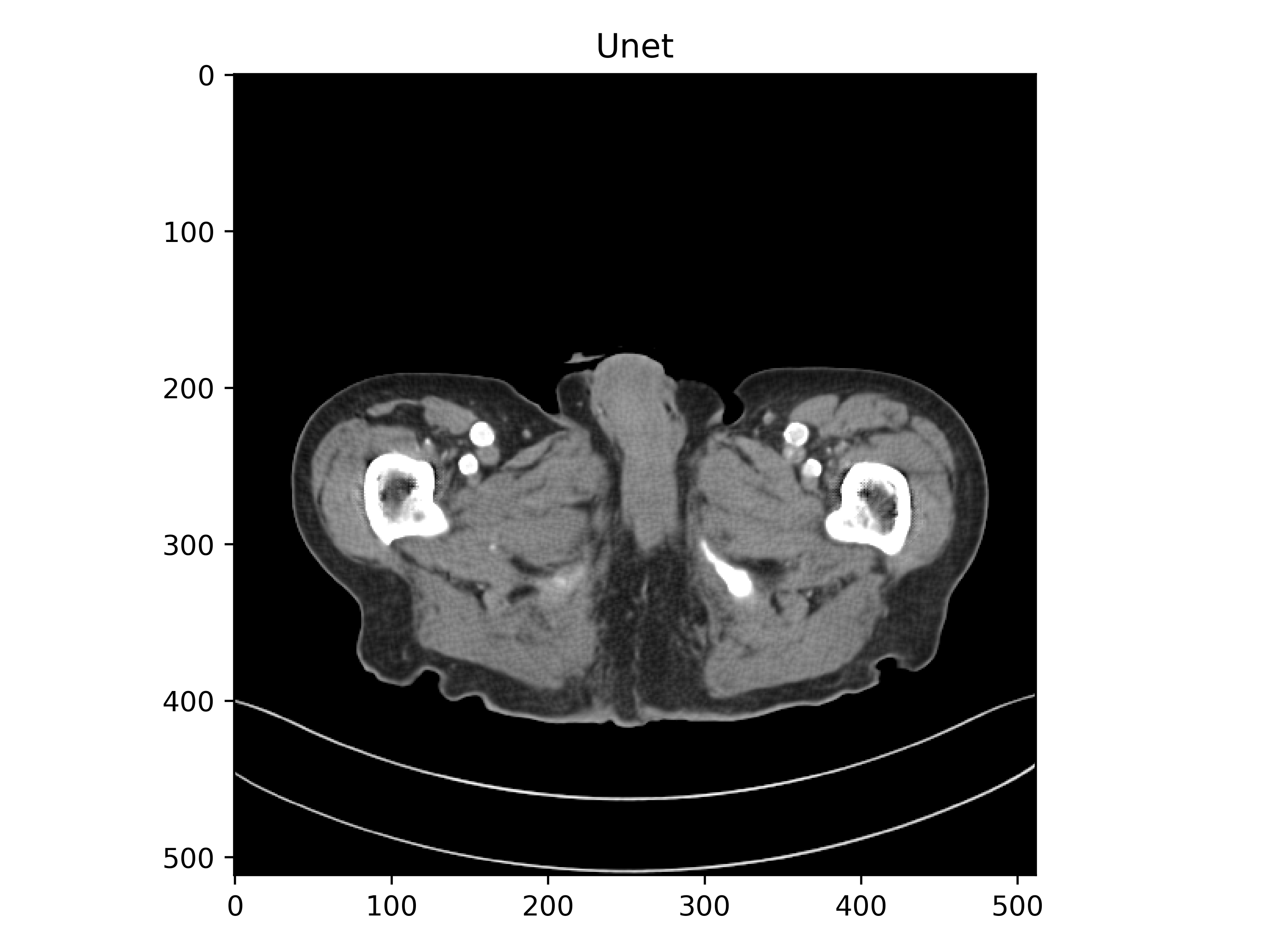}
	}
	\subfigure[]{
		\includegraphics[width=0.22\textwidth,trim=50 10 80 10, clip]{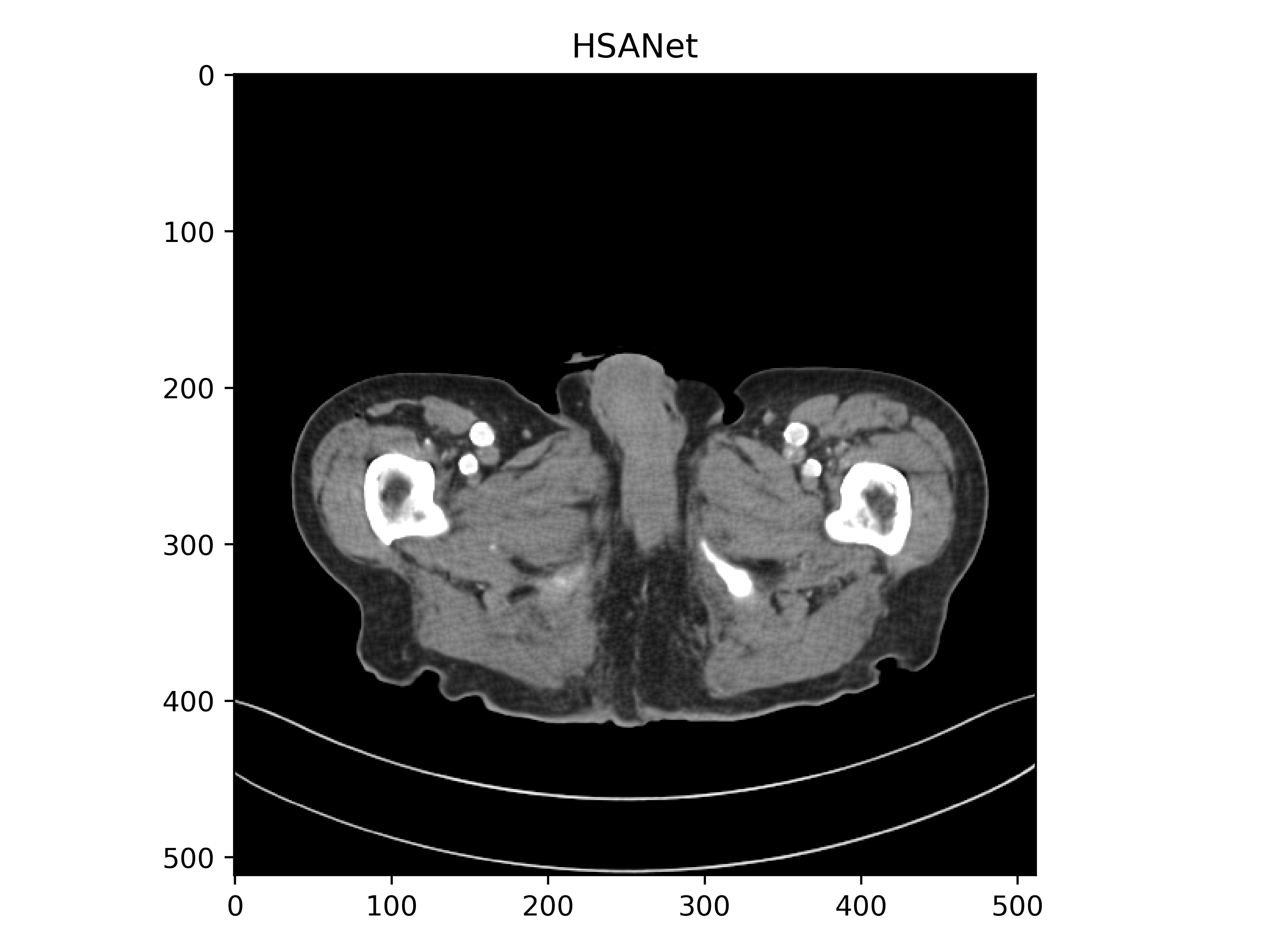}
	}
	\subfigure[]{
		\includegraphics[width=0.22\textwidth,trim=50 10 80 10, clip]{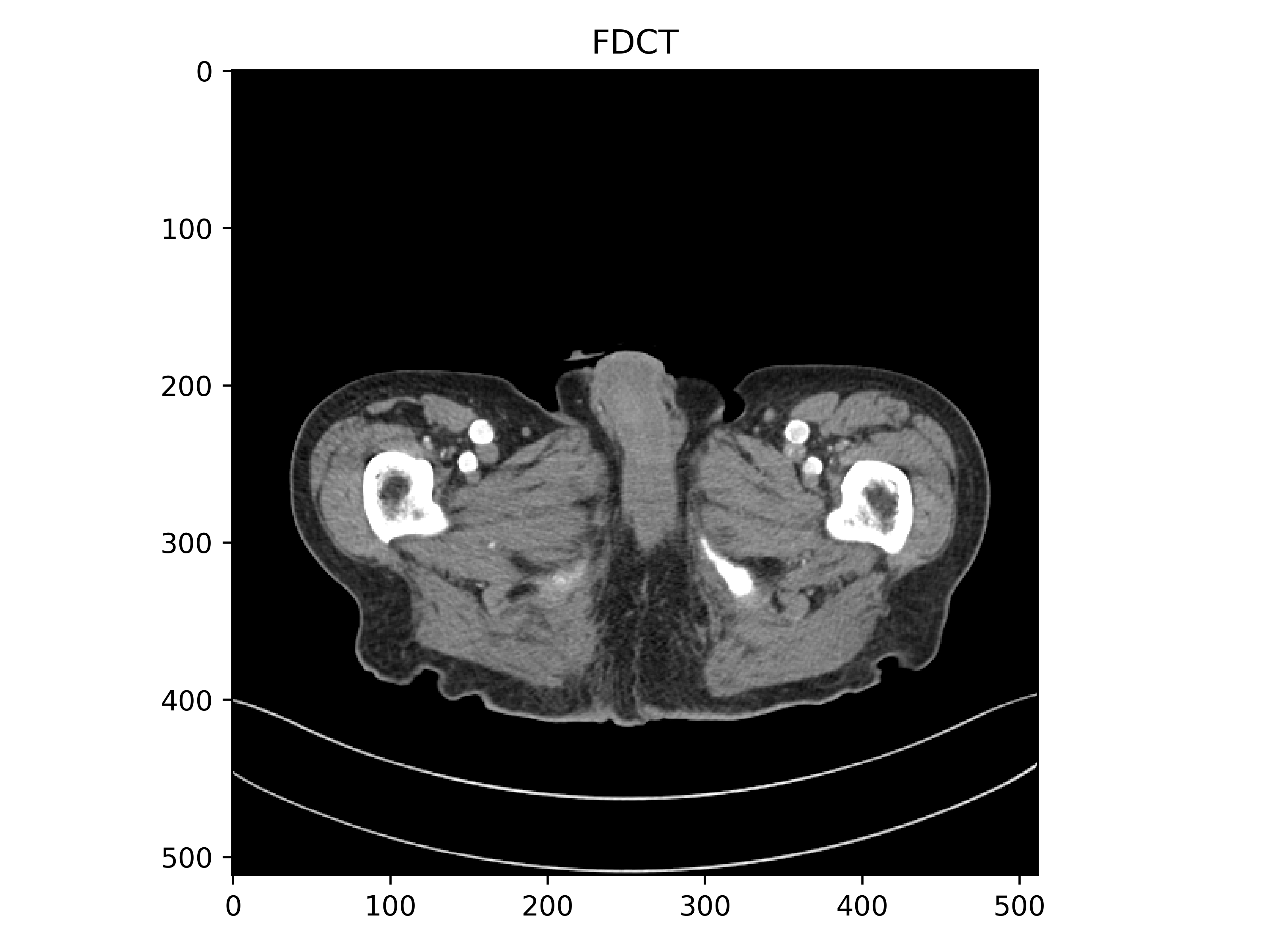}
	}
	
	\caption{Results of pelvis image for comparison. (a)LDCT, (b)RED-CNN,(c)Swin-Unet, (d)SwinIR, (e)CTformer, (f)Unet, (g)EHSANet, (h)FDCT}
	\label{LDCT_fig}
\end{figure*}

Although SwinIR is relatively stable and has a compact model size of 0.41M parameters, its denoising performance is insignificant (PSNR of 31.72), and it requires substantially more GPU memory—approximately 23 GB, which is four times higher than our method. Due to these substantial resource constraints, we were unable to evaluate larger variants of the SwinIR model for comparison. In contrast, our model not only delivers superior denoising accuracy but also demonstrates high stability. Additionally, it is considerably smaller than most competing models and consumes a similar amount of GPU memory, making it well-suited for practical deployment. Fig. \ref{LDCT_fig} presents a visual comparison of pelvis slices from a single patient in the LDCT test set. From the figure, we observe that RED-CNN, Swin-Unet, SwinIR, and CTformer effectively remove background noise, but at the cost of producing smoother images. In contrast, Unet reduces noise while preserving textural details, although some regions around the bone appear grid-like. EHSANet overcomes both limitations, suppressing noise while maintaining image sharpness and structural integrity, producing images that closely resemble FDCT.


\begin{table}[!ht]
\centering
\resizebox{\textwidth}{!}{%
\begin{tabular}{lcccccc}
\cline{1-7}
\multirow{2}{*}{Method}& \multirow{2}{*}{PSNR} &\multirow{2}{*}{SSIM} & \multirow{2}{*}{RMSE} & \multicolumn{1}{c}{GPU}  & \multirow{2}{*}{P}& \multirow{2}{*}{FLOPs}\\
& &&&{memory}&&\\
\cline{1-7} \\[-10pt]
LDCT & 28.84 & 0.86 & 15.16 & {-} &{-}&{-}\\
RED-CNN & 31.75$\pm$0.201 & 0.90$\pm 1.95e-3$ & 10.69$\pm$0.26&5.08G & 1.85M&4.66G\\
Swin-Unet & 31.06$\pm$0.047&0.89$\pm 3.72e-4$& 11.56$\pm$0.056& \bfseries1.2G& 0.95M&\bfseries0.04G\\
SwinIR & 31.72$\pm$0.011& 0.90$\pm 8.86e-5$ & 10.77$\pm$0.014& 23G &\bfseries0.41M&1.89G\\
CTformer&31.75$\pm$0.097&0.89$\pm 3.00e-3$&10.74$\pm$0.132&4.7G&1.45M&1.33G\\
Unet & 31.94$\pm$0.170 &0.90$\pm 7.69e-4$& 10.51$\pm$0.243&6G&31M&3.02G\\
ours&\bfseries32.21$\pm$0.029 & \bfseries0.90$\pm \mathbf{5.90e-4}$&\bfseries10.30$\pm$0.040 &5.4G&0.61M&0.37G\\
\cline{1-7}
\end{tabular}}
\caption{Quantitative evaluation for LDCT dataset over multiple training runs. P represents for network parameters}\label{comparison}
\end{table}

\subsubsection{Quantitative evaluation results on LDPET dataset}
\begin{figure*}[!h]
	\centering
	\subfigure[]{
		\includegraphics[width=0.3\textwidth,trim=0 0 0 0, clip]{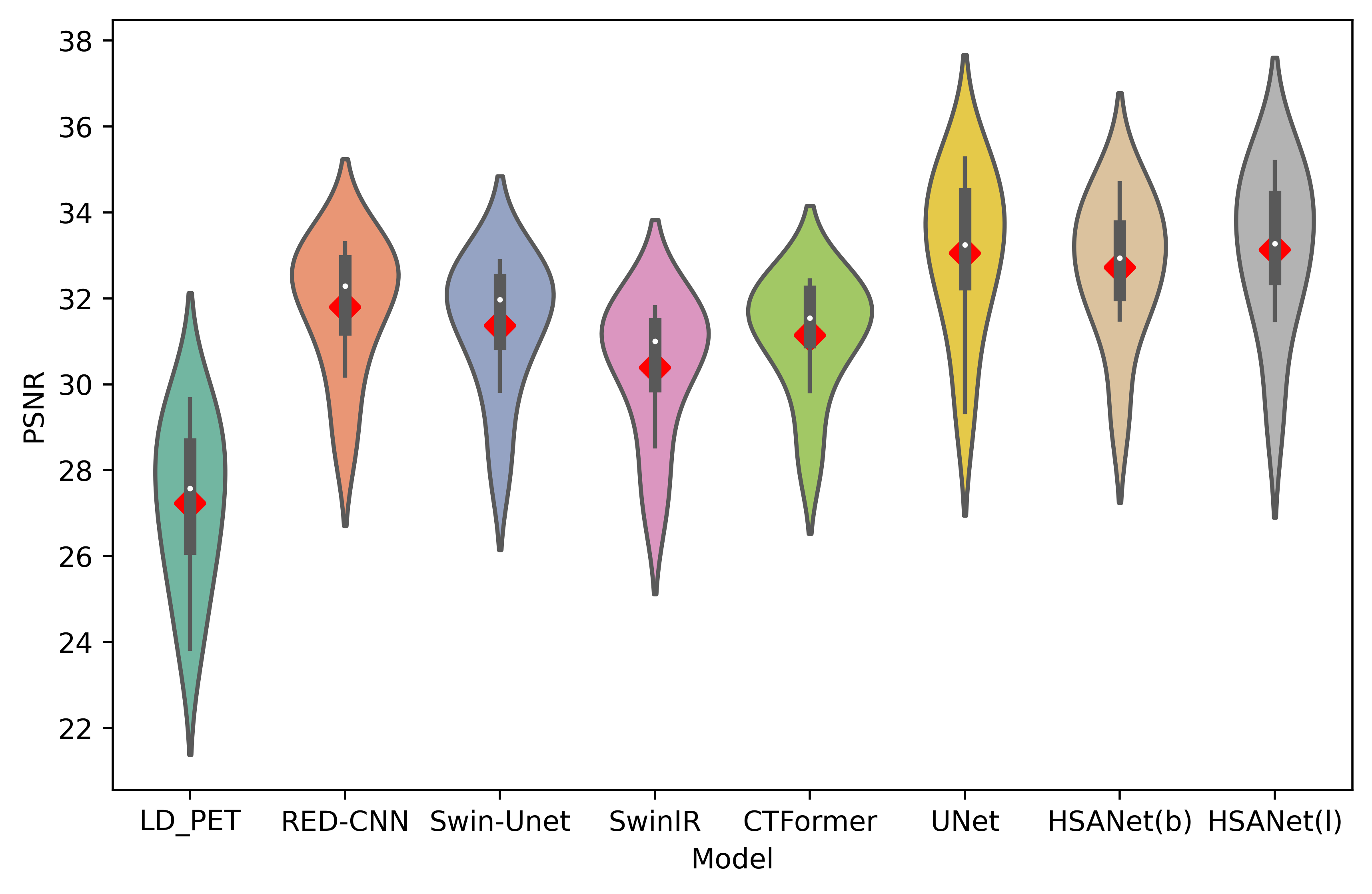}
	}
	\subfigure[]{
		\includegraphics[width=0.3\textwidth,trim=0 0 0 0, clip]{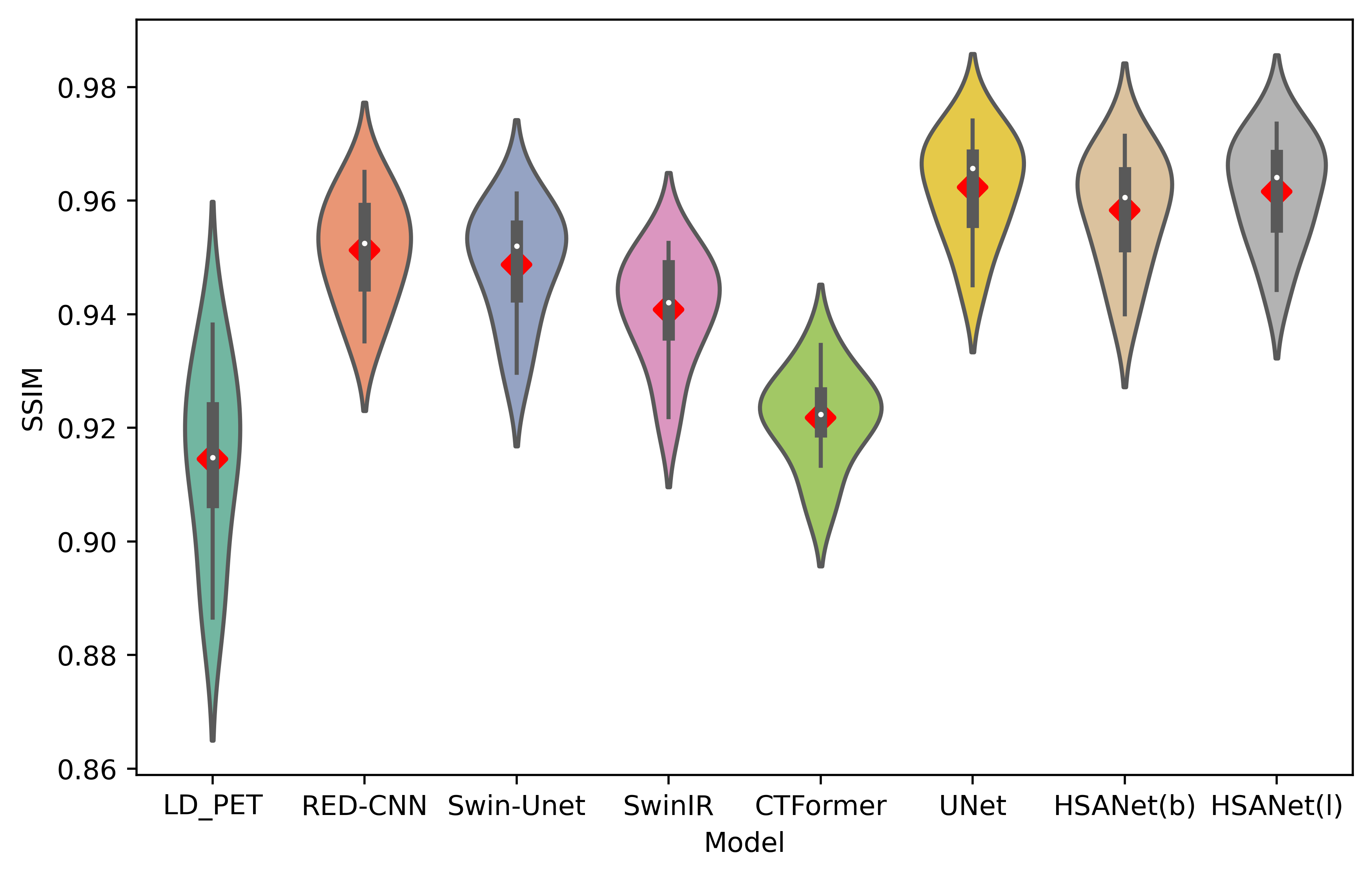}
	}
	\subfigure[]{
		\includegraphics[width=0.3\textwidth,trim=0 0 0 0, clip]{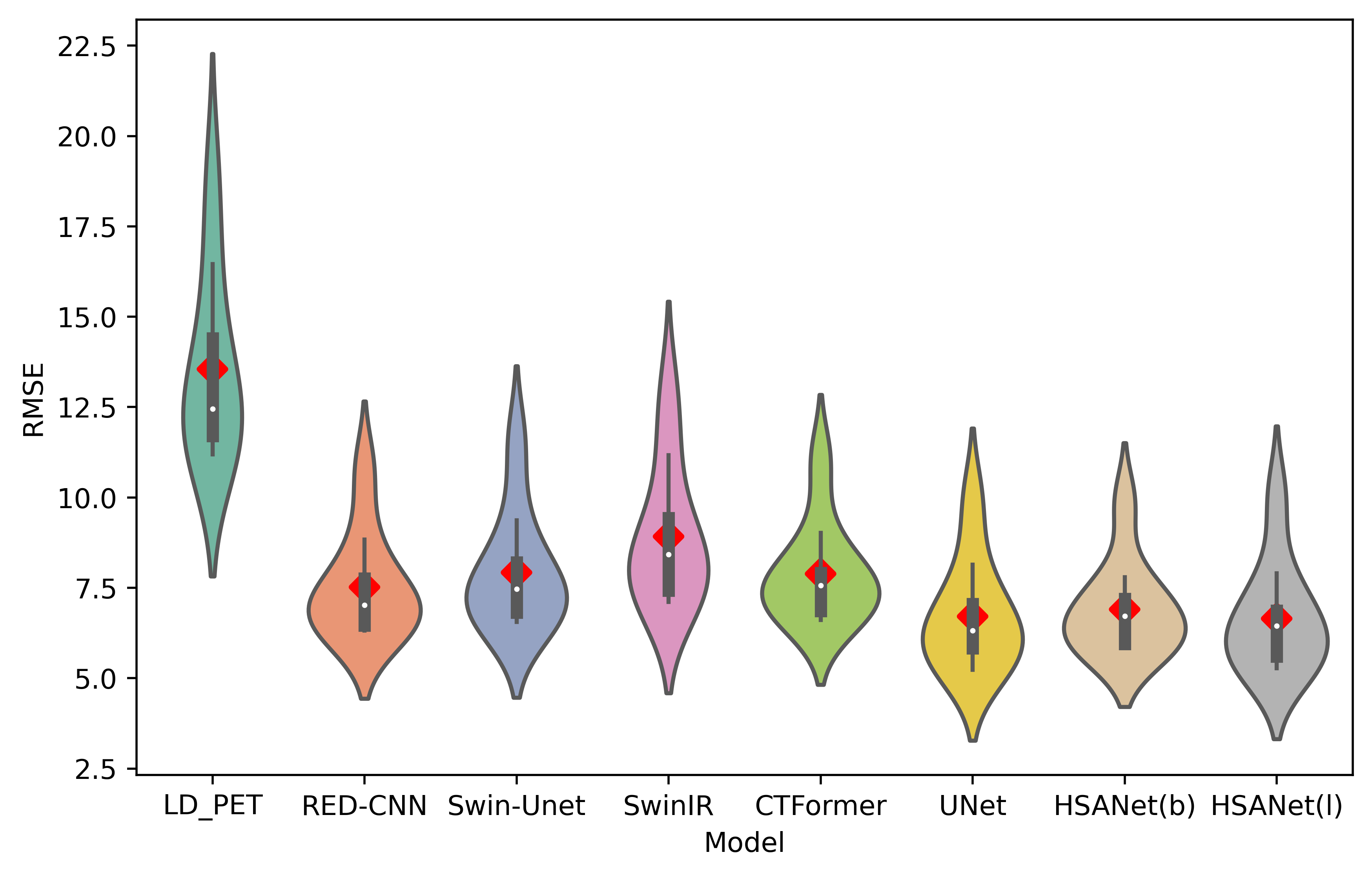}
	}
	\caption{Quantitative (a)PSNR, (b)SSIM and (c)RMSE of 9 different PET patients on test set. Red points are average. Width of violin plot represent the density of data at each value. EHSANet(b) and EHSANet(l) represent base and large model respectively. Quartiles are shown as thick lines inside the violin plot}
	\label{LDPET_violin}
\end{figure*}

We have used the max-min normalization method to normalize images for training. An average test results on all 9 patients are shown in Table \ref{comparison_pet}. As shown in the table, our base model outperforms the other approaches and achieves a PSNR of 32.73, except Unet. Although its performance is 0.31 dB lower than Unet, it achieves a comparable SSIM with substantially fewer parameters. The model was further scaled up by increasing the number of residual convolutional layers with skip-connections and incorporating two residual Swin-Transformer bottleneck blocks, each containing four bottleneck units. In addition, since the ESGA module within the Swin-Transformer block did not improve performance, the original Swin-Transformer module was retained. Fig. \ref{LDPET_violin} shows the violin plots of the scaled large EHSANet for the 9 patients in the test set. Our model demonstrates slightly better results on PSNR and RMSE performance across patients compared with Unet.

\begin{figure*}[!h]
	\centering
	\subfigure[]{
		\includegraphics[width=0.22\textwidth,trim=50 10 80 10, clip]{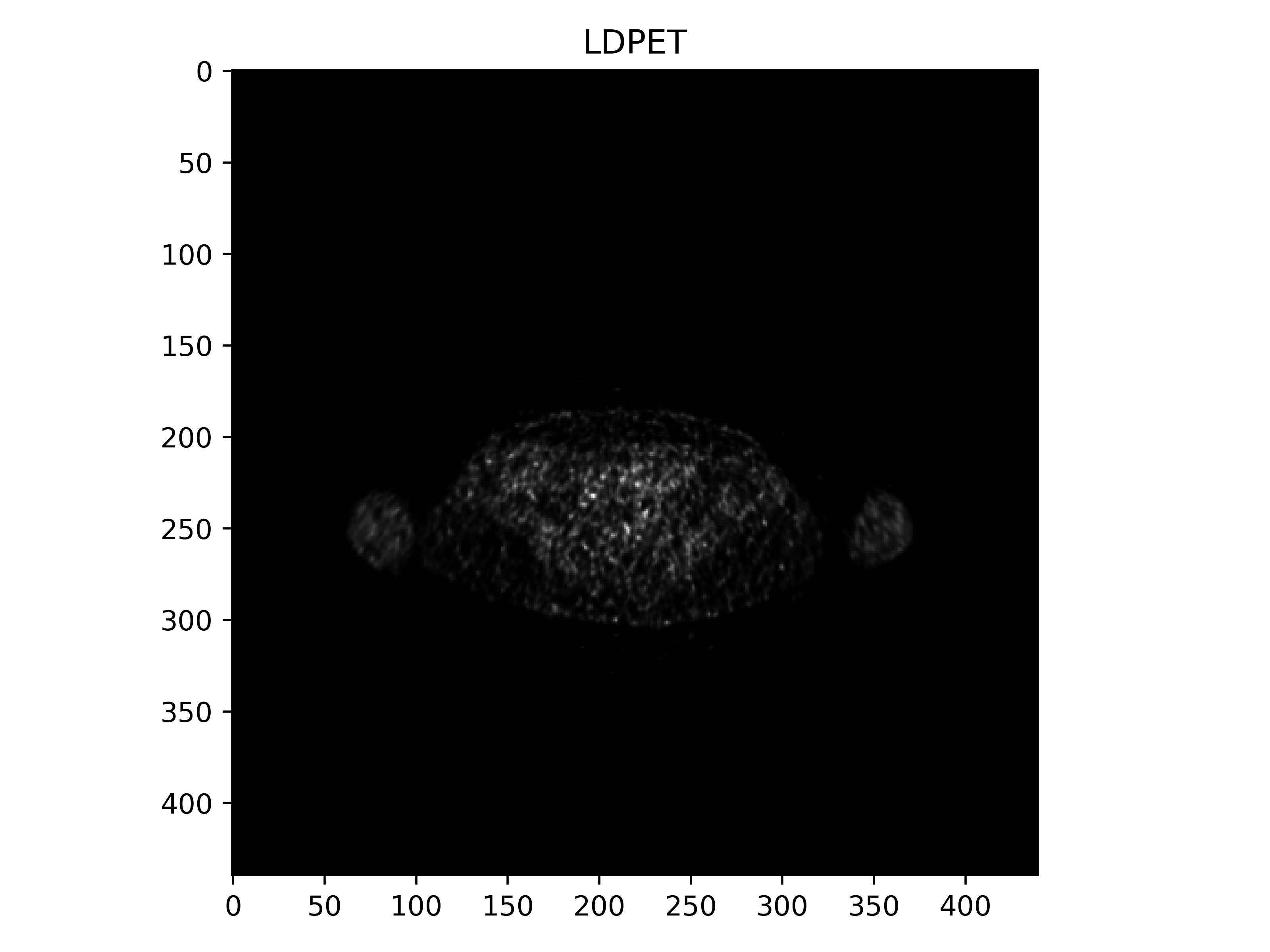}
	}
	\subfigure[]{
		\includegraphics[width=0.22\textwidth,trim=50 10 80 10, clip]{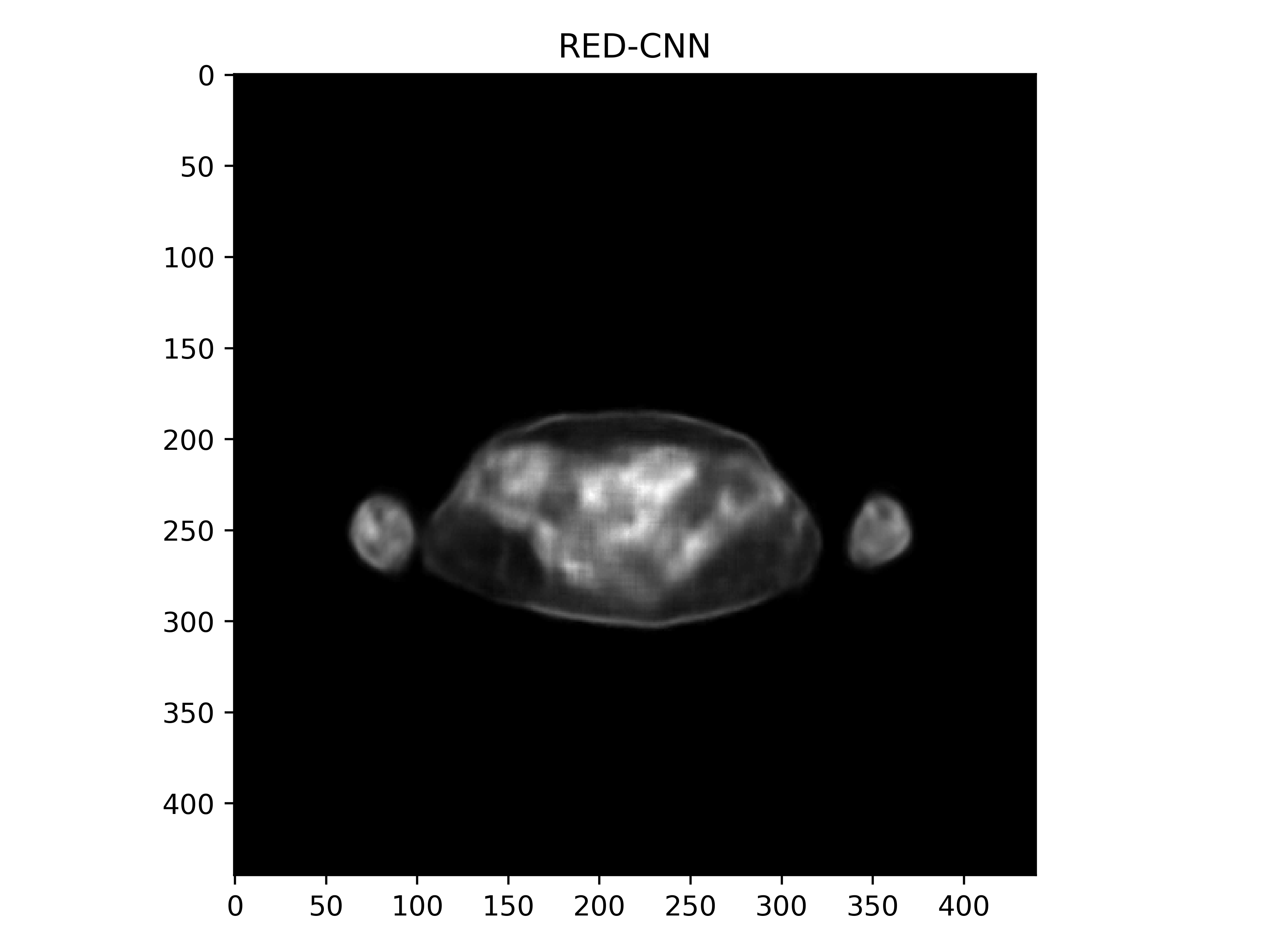}
	}
	\subfigure[]{
		\includegraphics[width=0.22\textwidth,trim=50 10 80 10, clip]{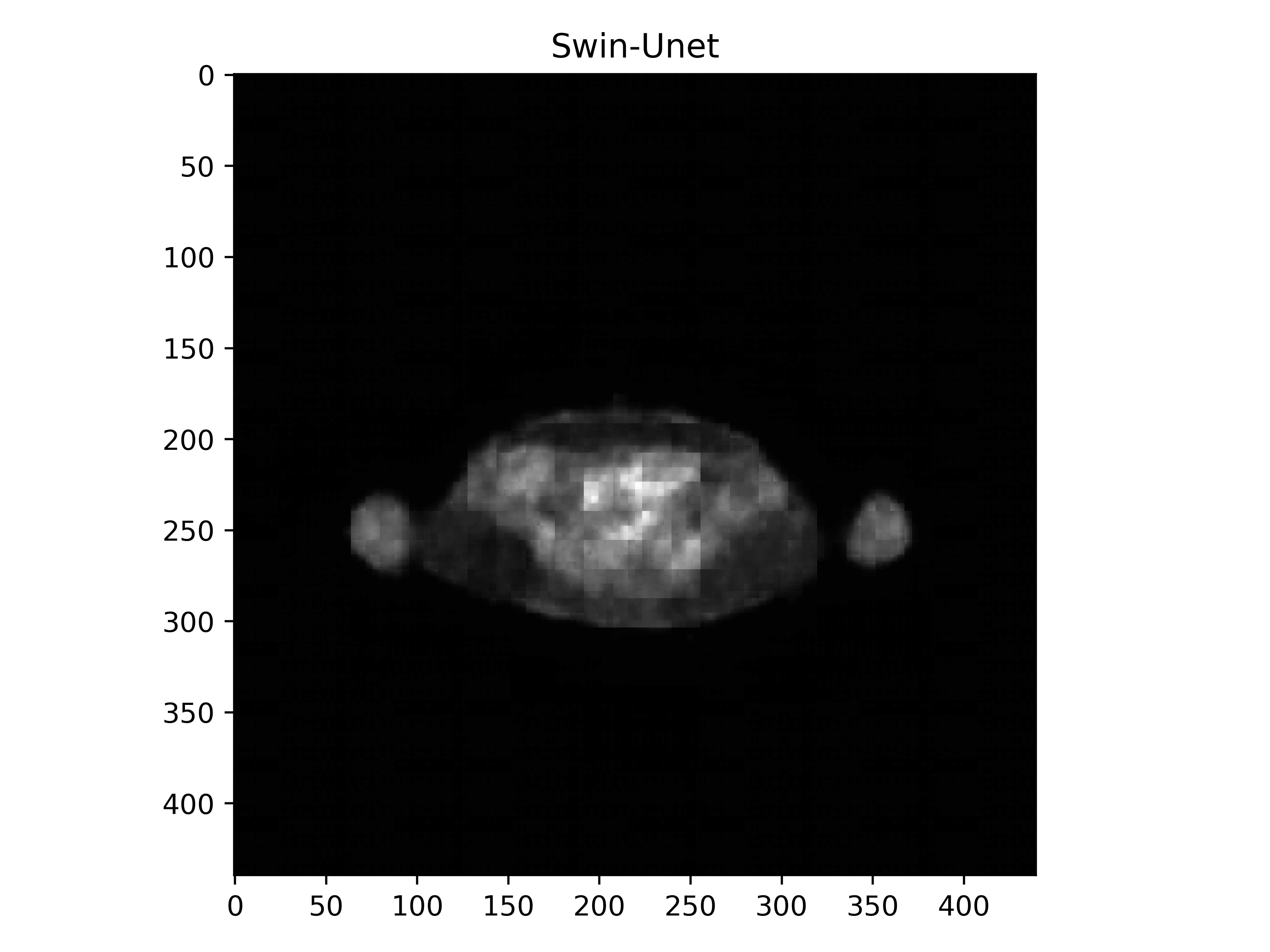}
	}
	\subfigure[]{
		\includegraphics[width=0.22\textwidth,trim=50 10 80 10, clip]{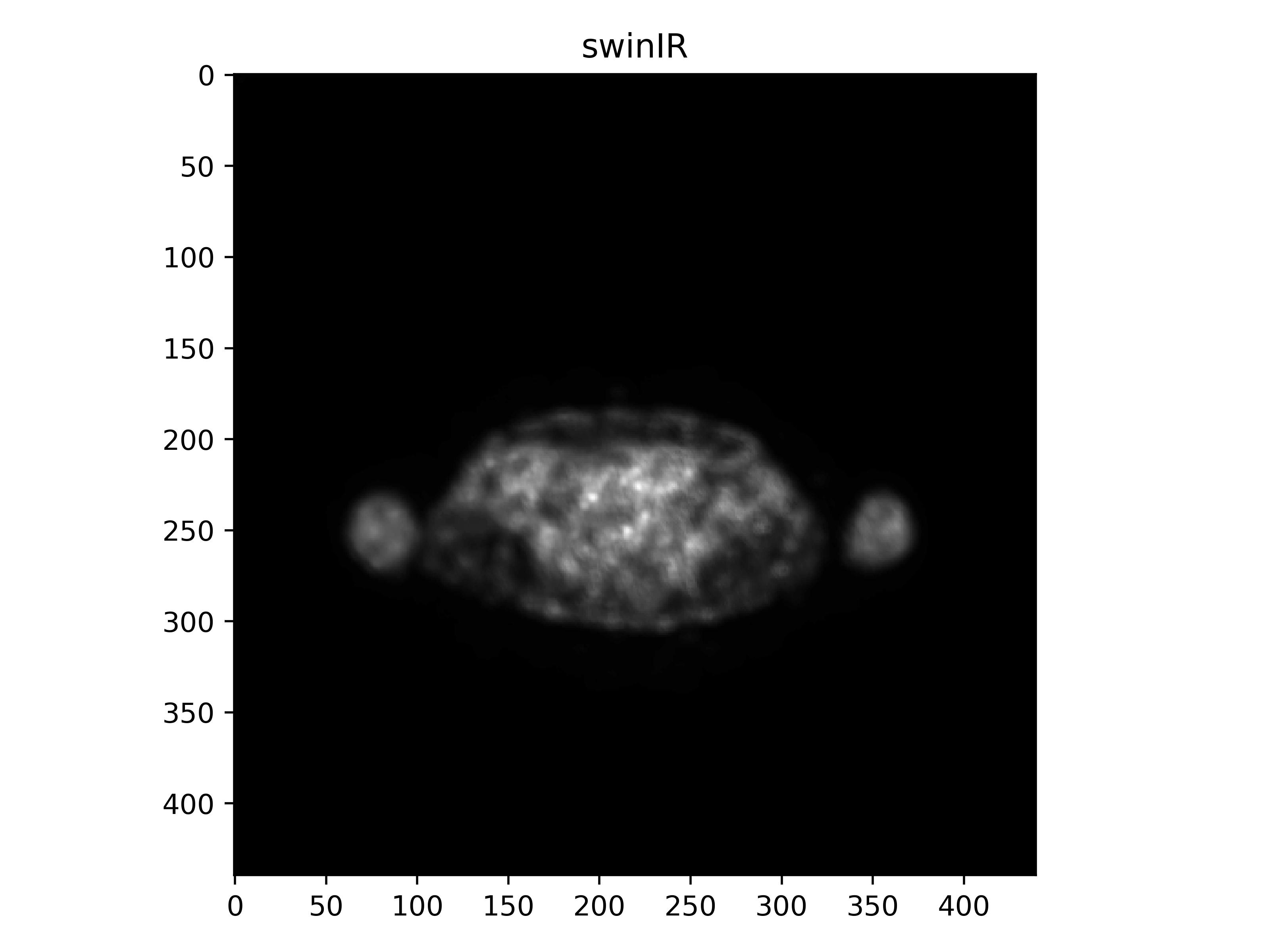}
	}
	\quad
	\subfigure[]{
		\includegraphics[width=0.22\textwidth,trim=50 10 80 10, clip]{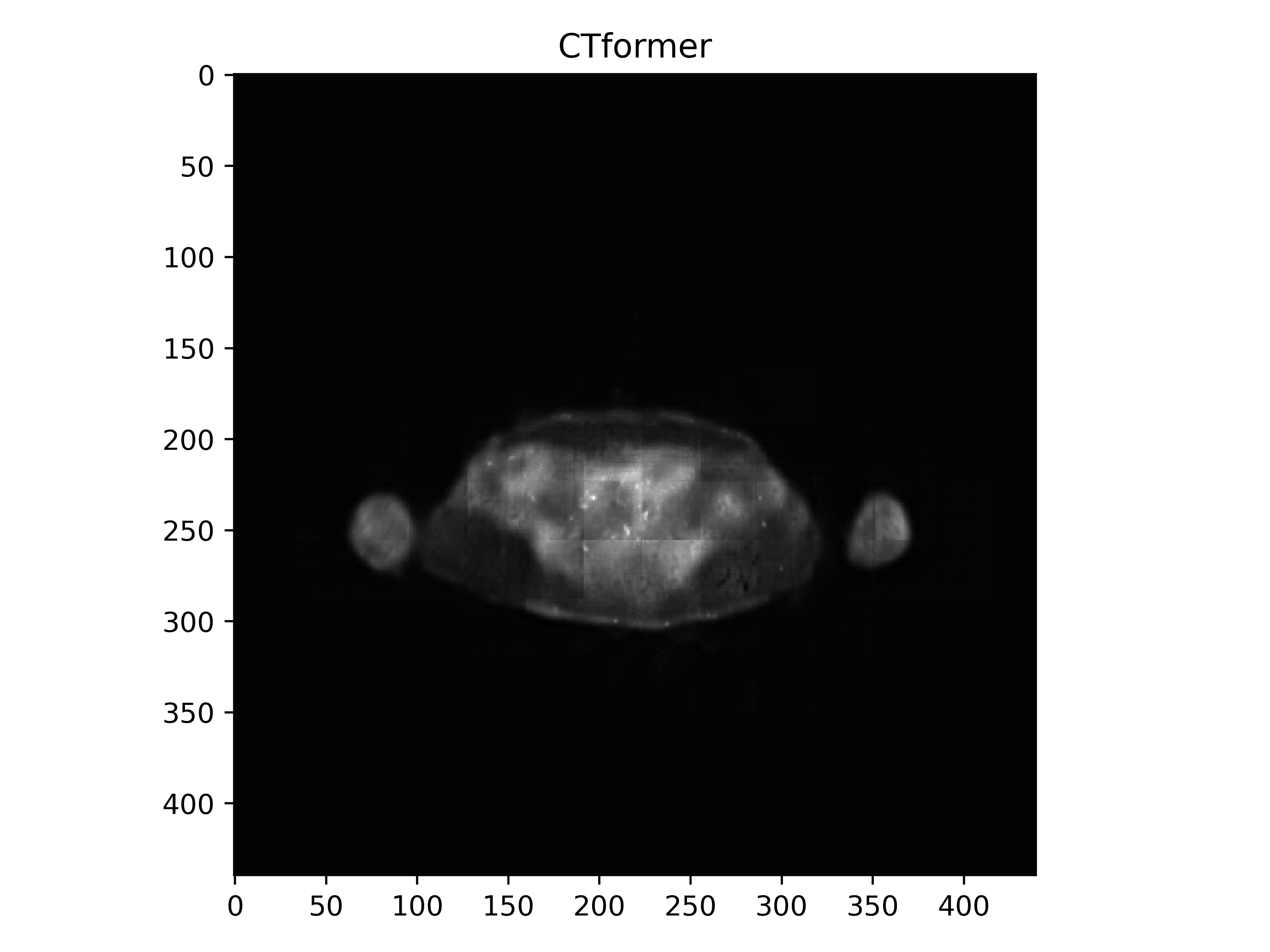}
	}
	\subfigure[]{
		\includegraphics[width=0.22\textwidth,trim=50 10 80 10, clip]{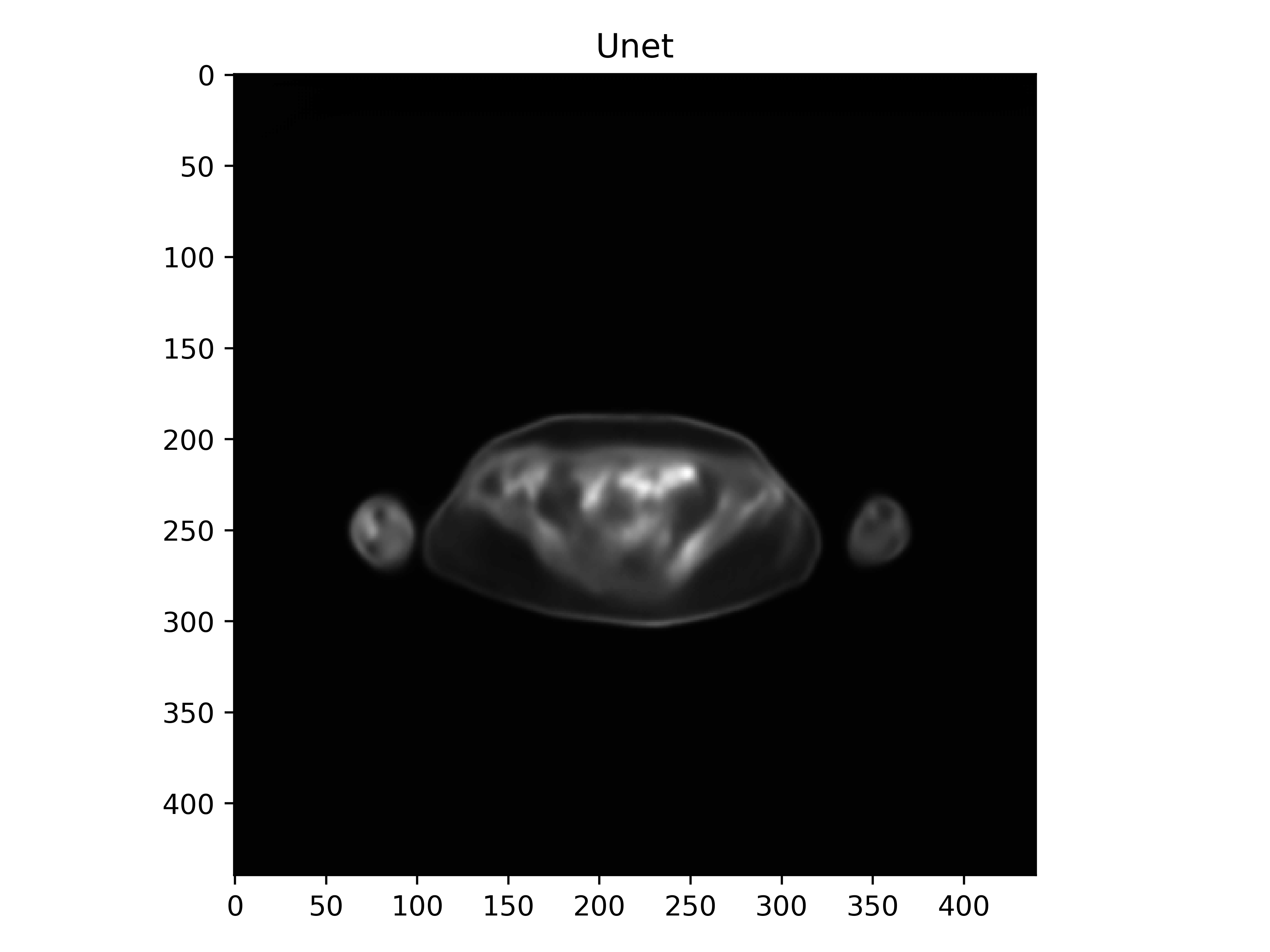}
	}
	\subfigure[]{
		\includegraphics[width=0.22\textwidth,trim=50 10 80 10, clip]{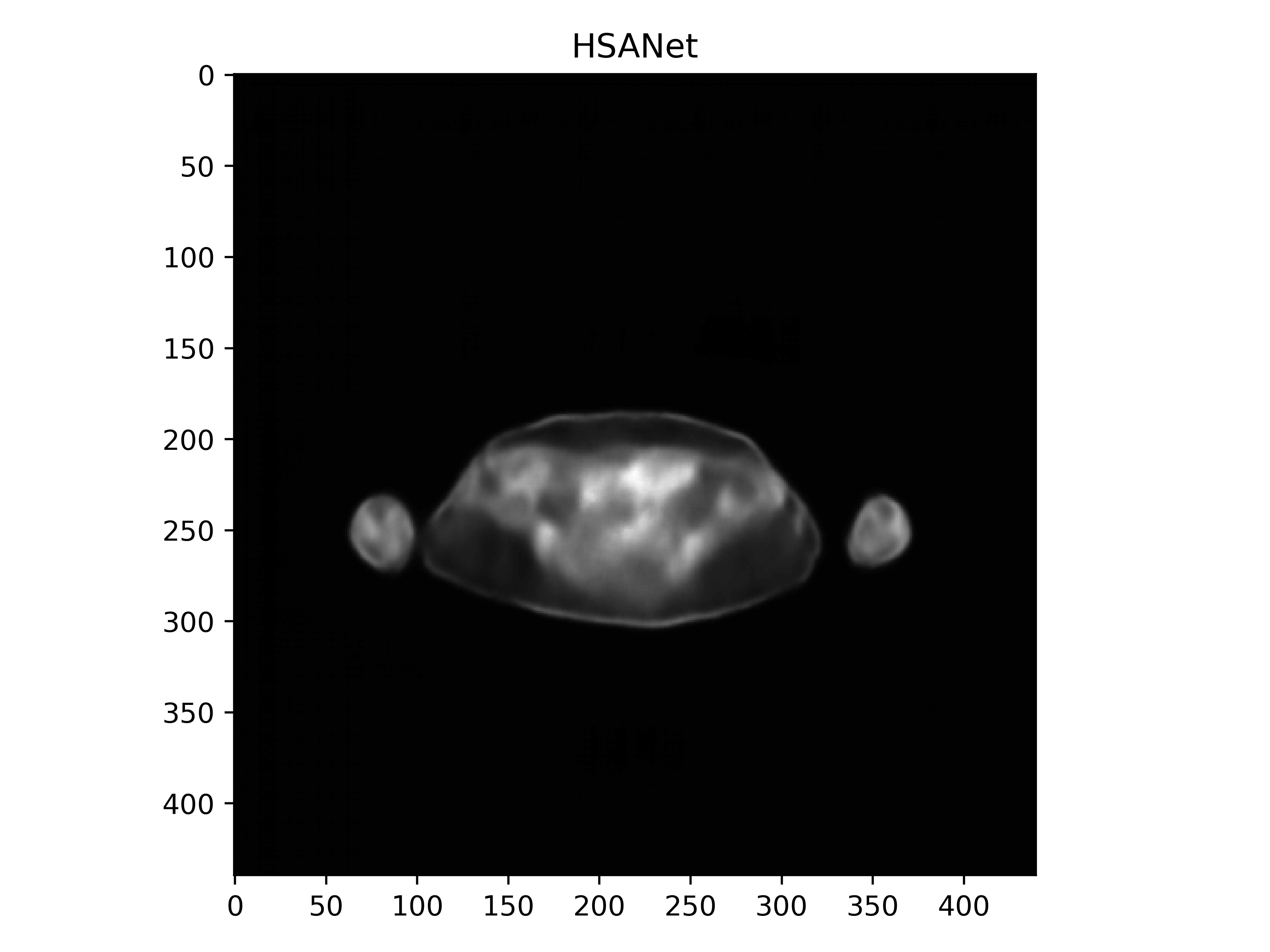}
	}
	\subfigure[]{
		\includegraphics[width=0.22\textwidth,trim=50 10 80 10, clip]{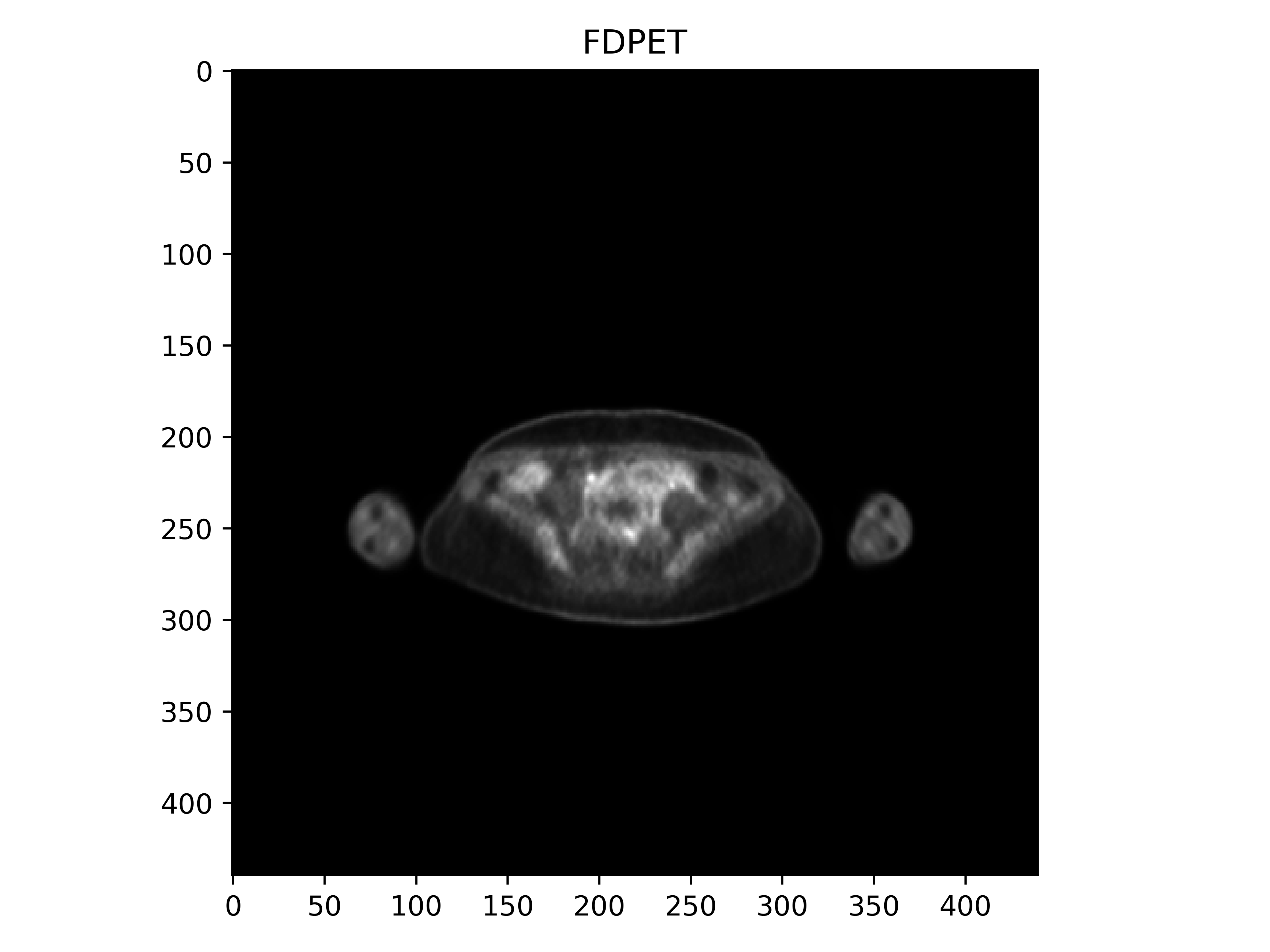}
	}
	
	\caption{Results of abdomen image for comparison. (a)LDPET, (b)RED-CNN,(c)Swin-Unet, (d)SwinIR, (e)CTformer, (f)Unet, (g)EHSANet large, (h)FDPET}
	\label{LDPET_fig}
\end{figure*}

Fig. \ref{LDPET_fig} shows the denoised results of an abdominal image. As observed, CTformer and Swin-Unet produce grid-like artifacts in the image. RED-CNN and SwinIR fail to capture the image patterns. U-Net and EHSANet appear to capture them better than the other methods, although their results are overly smooth.

\begin{table}[!h]
\centering
\begin{tabular}{lccccc}
\cline{1-6}
Methods & PSNR &SSIM & RMSE & GPU memory & parameters \\
\cline{1-6}
LDPET & 27.23 & 0.91 & 13.55 & - & - \\
RED-CNN & 31.79 & 0.951 & 7.52 & 5.08G & 1.85M \\
Swin-Unet & 31.36 & 0.95 & 7.92 & \textbf{1.2G} & 0.95M \\
SwinIR & 30.39 & 0.94 & 8.9 & 23G & \textbf{0.41M} \\
CTformer & 31.14 & 0.92 & 7.88 & 4.7G & 1.45M \\
Unet & 33.04 & 0.96 & 6.72 & 6G & 31M \\
EHSANet(base) & 32.73 & 0.96 & 6.9 & 5.4G & 0.61M \\
EHSANet(large) & \textbf{33.11} & \textbf{0.96} & \textbf{6.55} & 10.4G & 3.88M \\
\cline{1-6}
\end{tabular}
\caption{Quantitative evaluation for LDPET dataset.}\label{comparison_pet}
\end{table}

\subsubsection{Quantitative evaluation results on LDCT chest dataset}
\begin{figure*}[!h]
	\centering
	\subfigure[]{
		\includegraphics[width=0.3\textwidth,trim=0 0 0 0, clip]{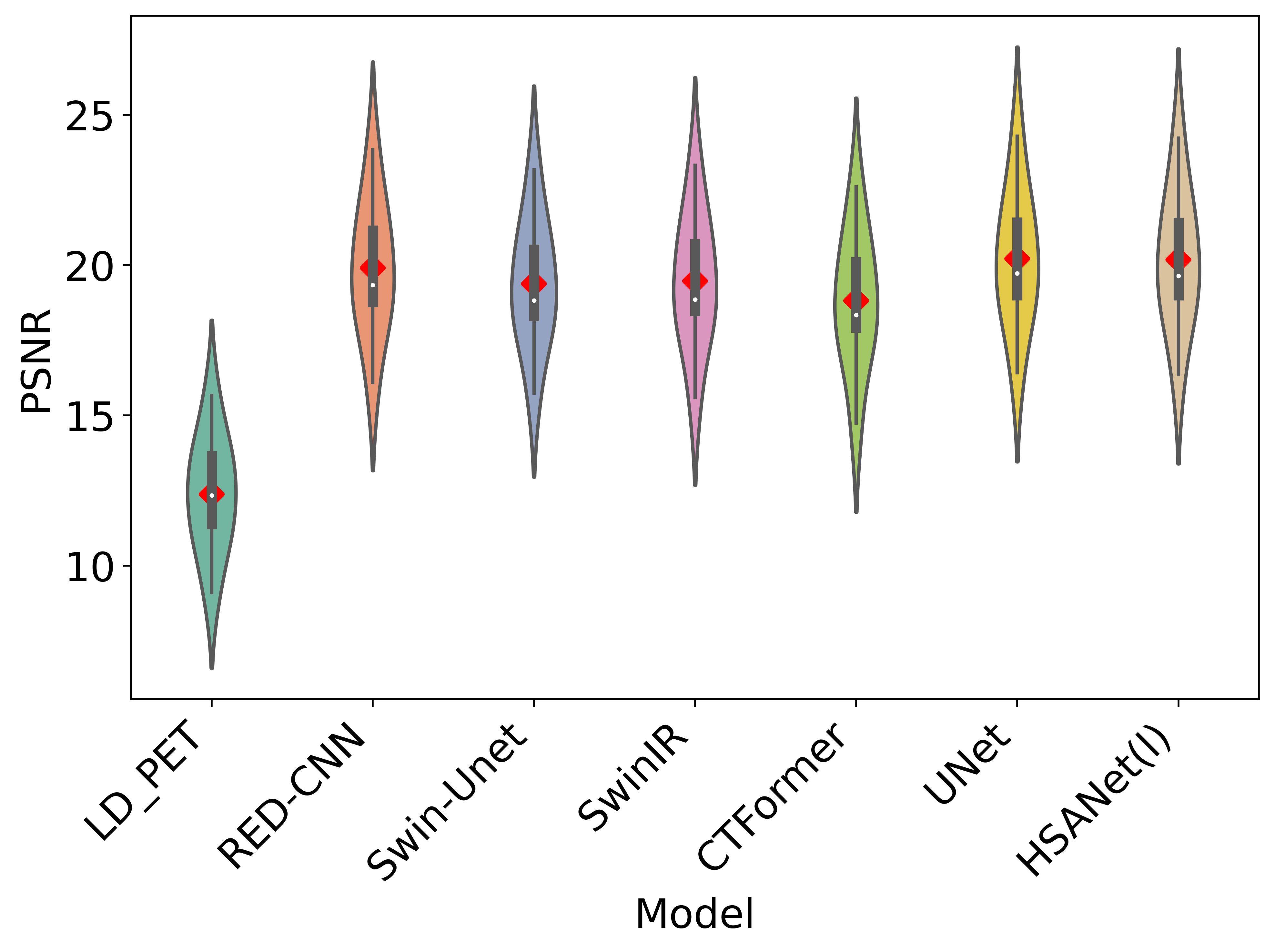}
	}
	\subfigure[]{
		\includegraphics[width=0.3\textwidth,trim=0 0 0 0, clip]{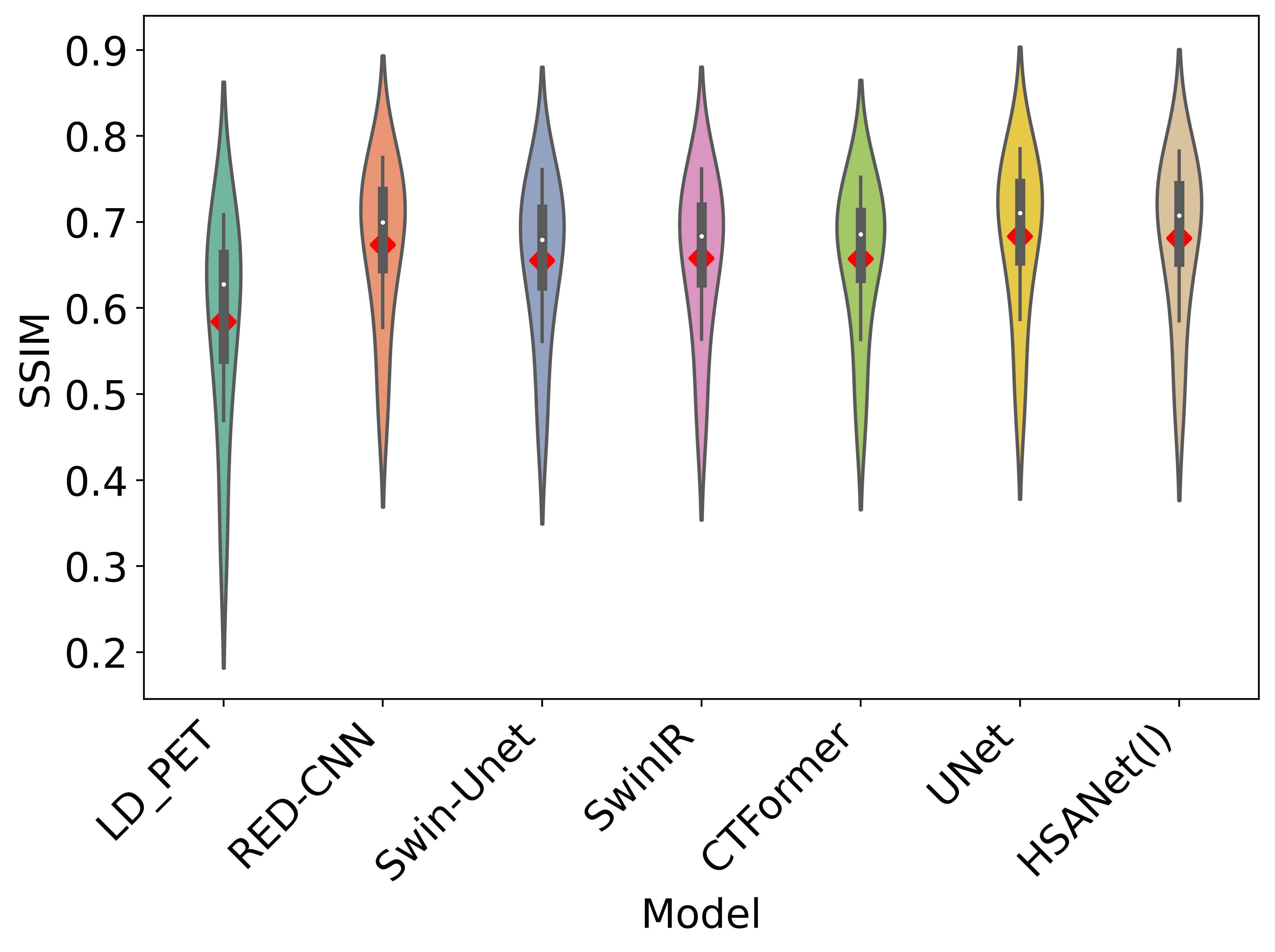}
	}
	\subfigure[]{
		\includegraphics[width=0.3\textwidth,trim=0 0 0 0, clip]{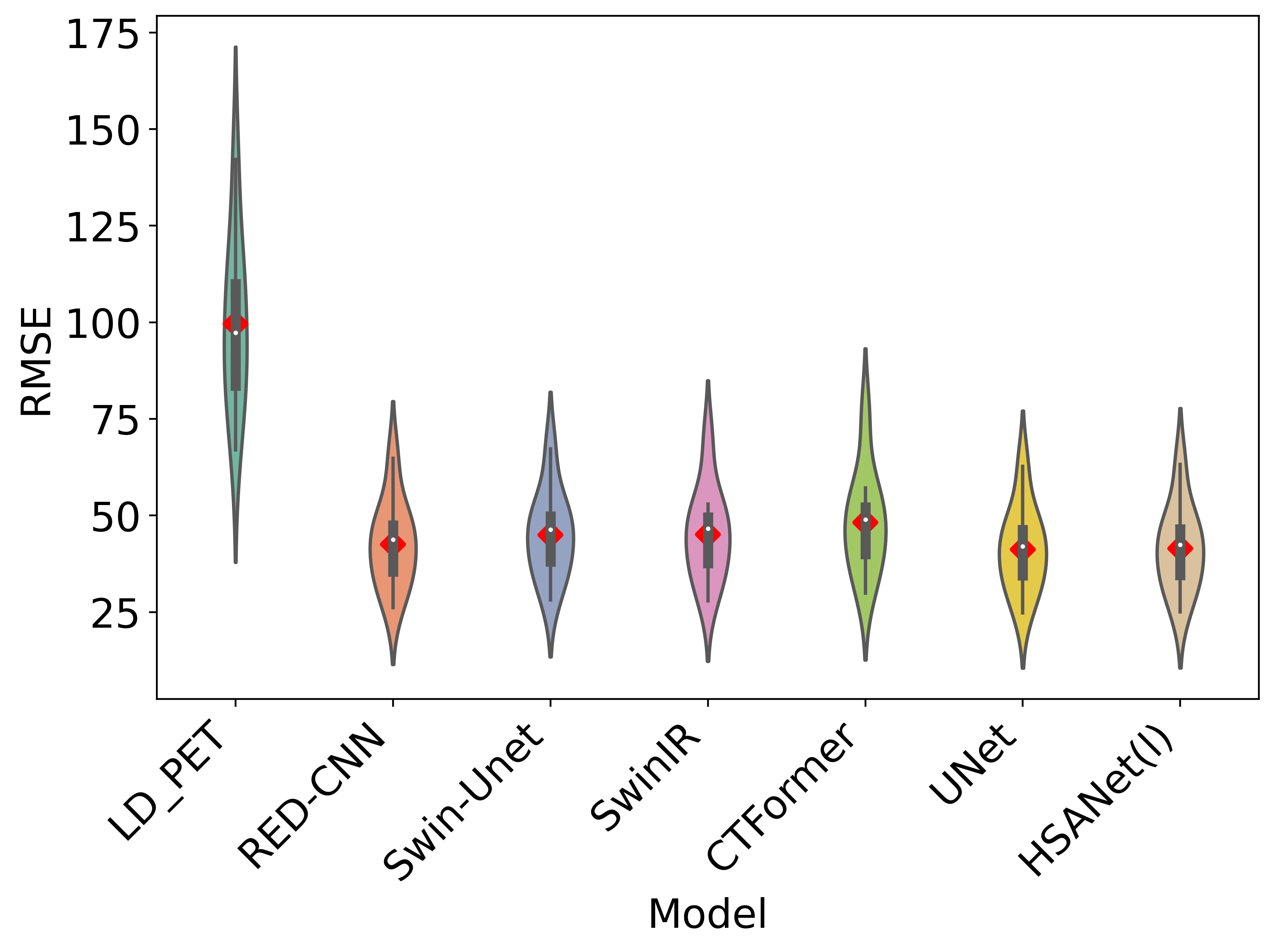}
	}
	\caption{Quantitative (a)PSNR, (b)SSIM and (c)RMSE of 9 different mayo chest LDCT patients on test set. Red points are average. Width of violin plot represent the density of data at each value.  EHSANet(l) represents large EHSANet model. Quartiles are shown as thick lines inside the violin plot}
	\label{LDCT_chest_violin}
\end{figure*}

\begin{figure*}[!h]
	\centering
	\subfigure[]{
		\includegraphics[width=0.22\textwidth,trim=50 10 80 10, clip]{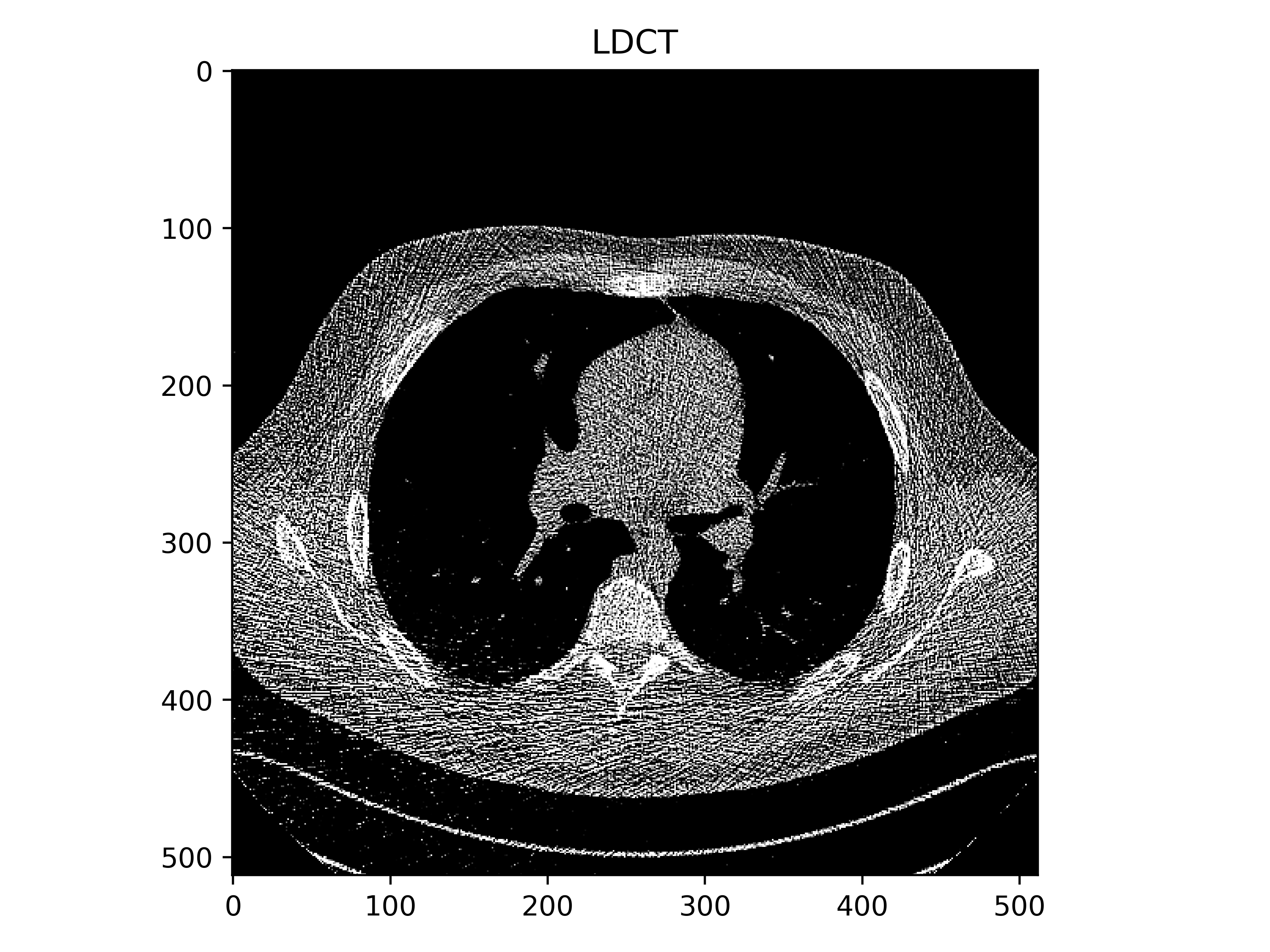}
	}
	\subfigure[]{
		\includegraphics[width=0.22\textwidth,trim=50 10 80 10, clip]{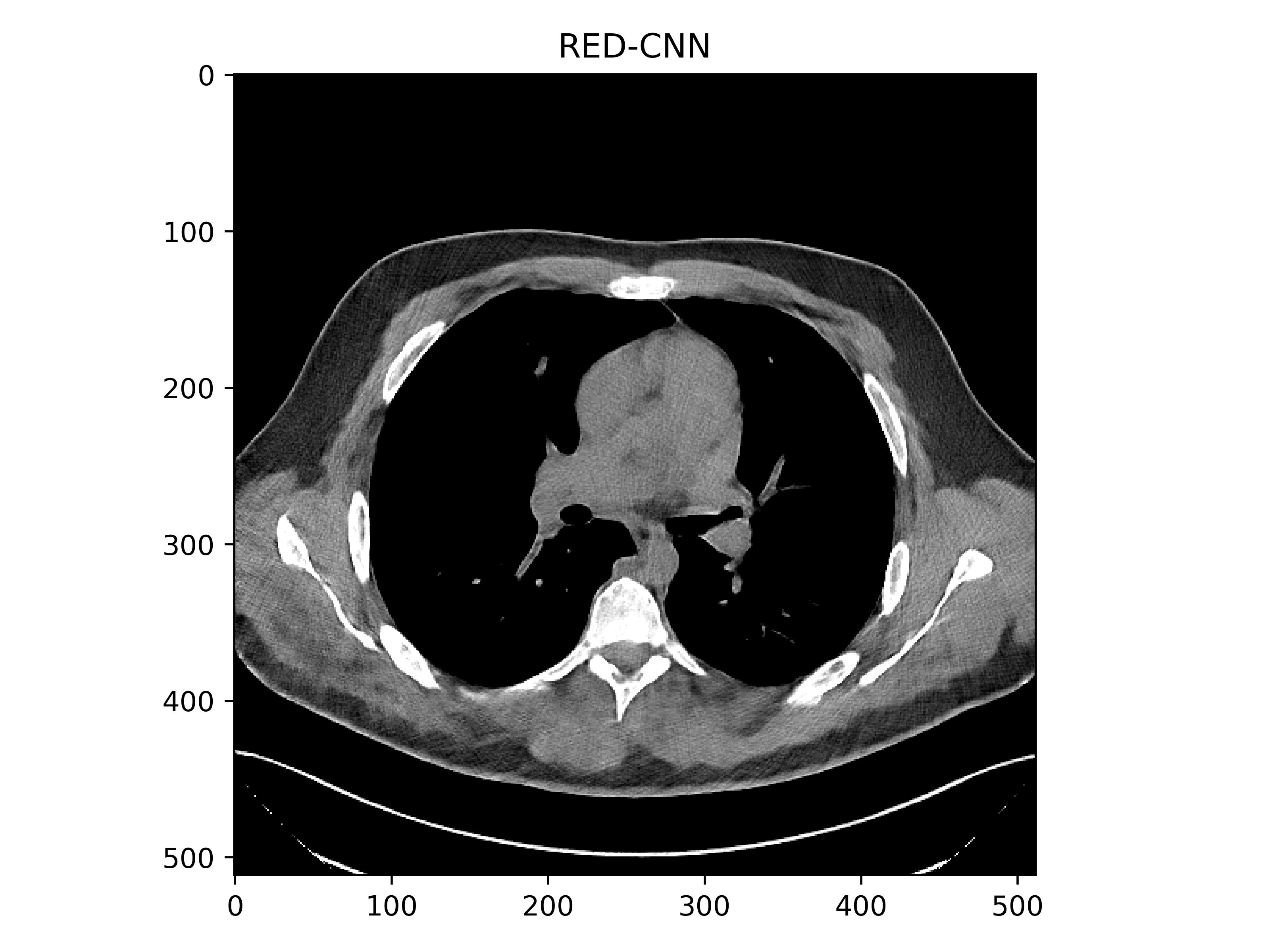}
	}
	\subfigure[]{
		\includegraphics[width=0.22\textwidth,trim=50 10 80 10, clip]{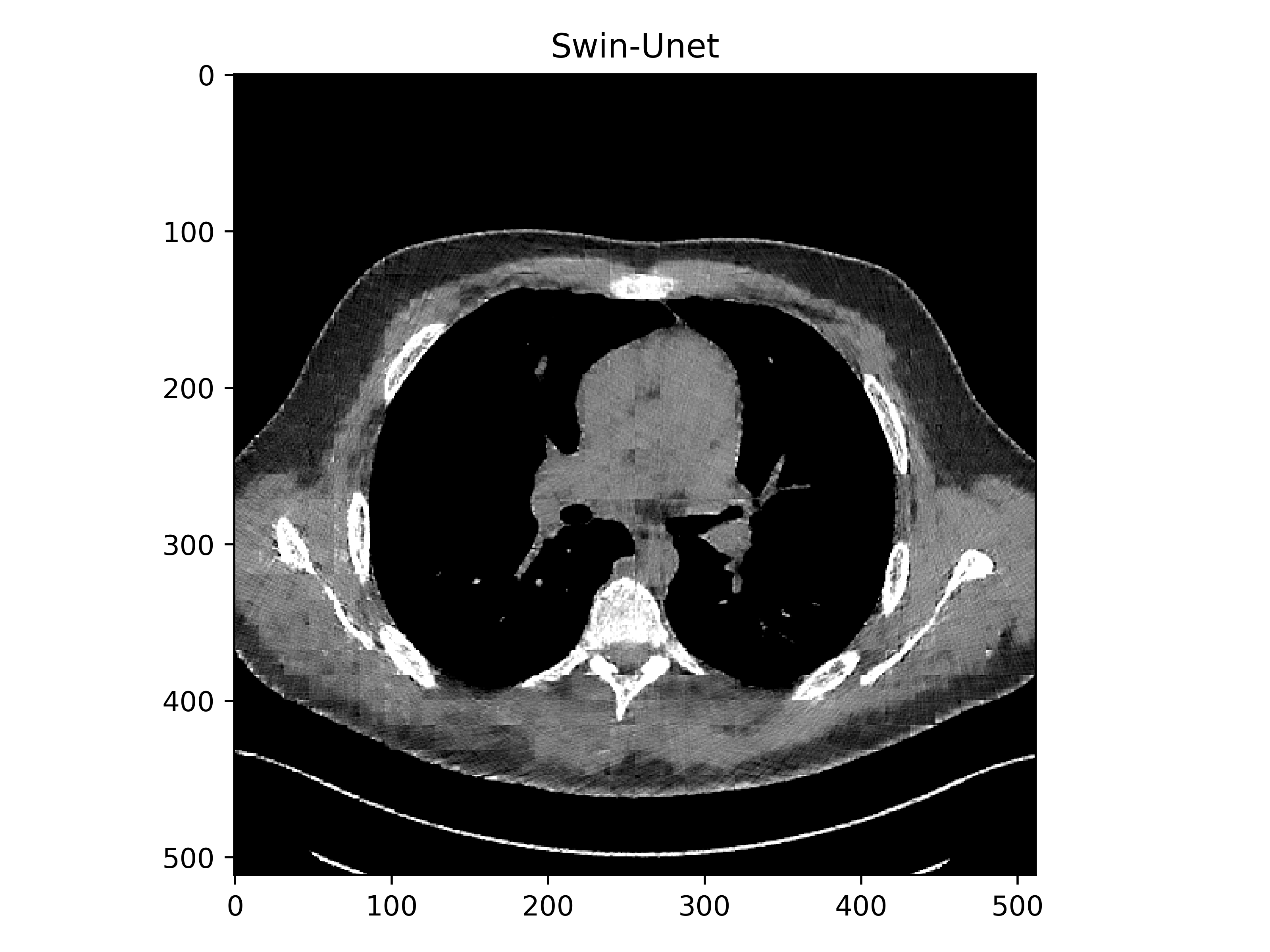}
	}
	\subfigure[]{
		\includegraphics[width=0.22\textwidth,trim=50 10 80 10, clip]{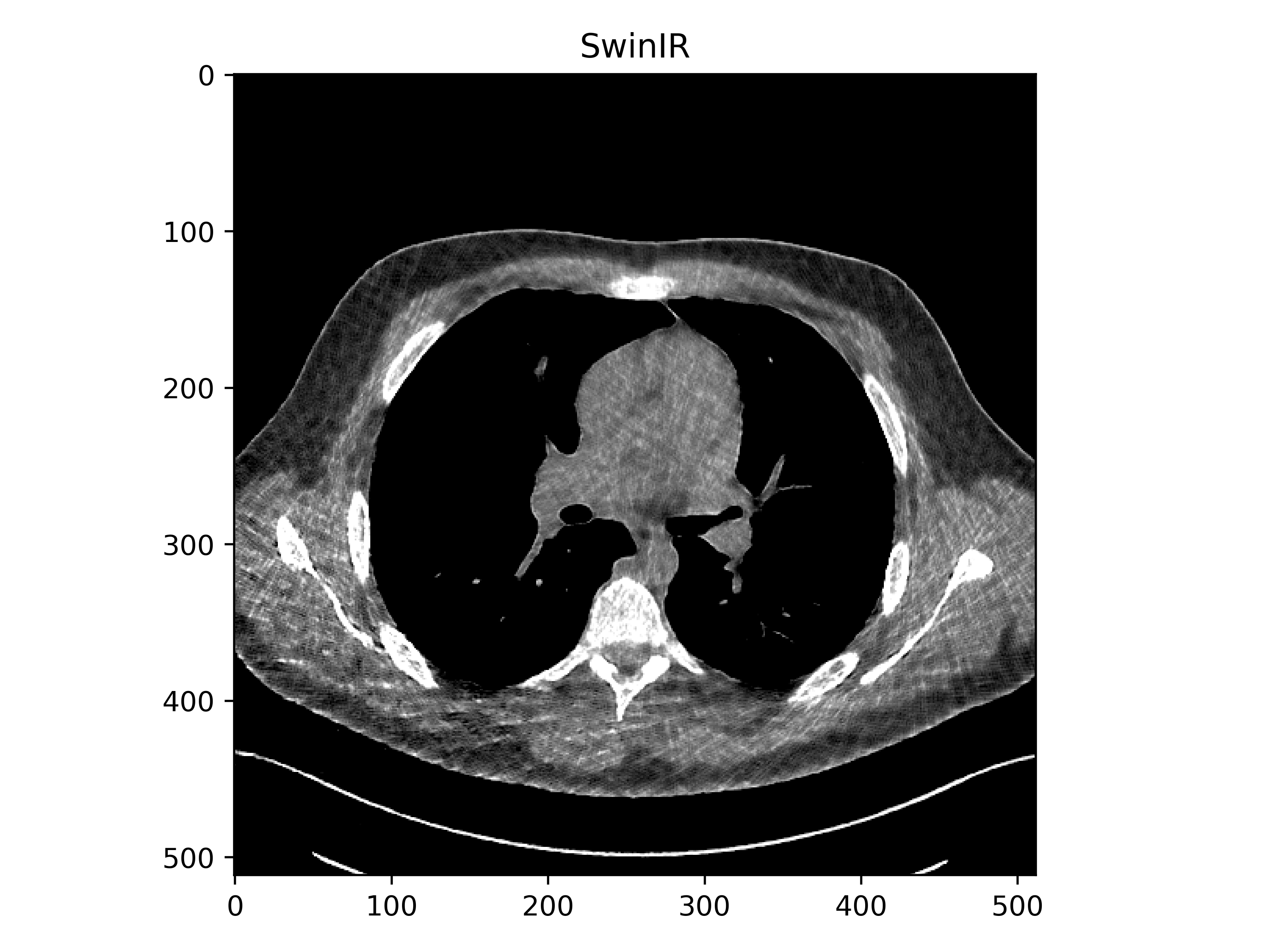}
	}
	\quad
	\subfigure[]{
		\includegraphics[width=0.22\textwidth,trim=50 10 80 10, clip]{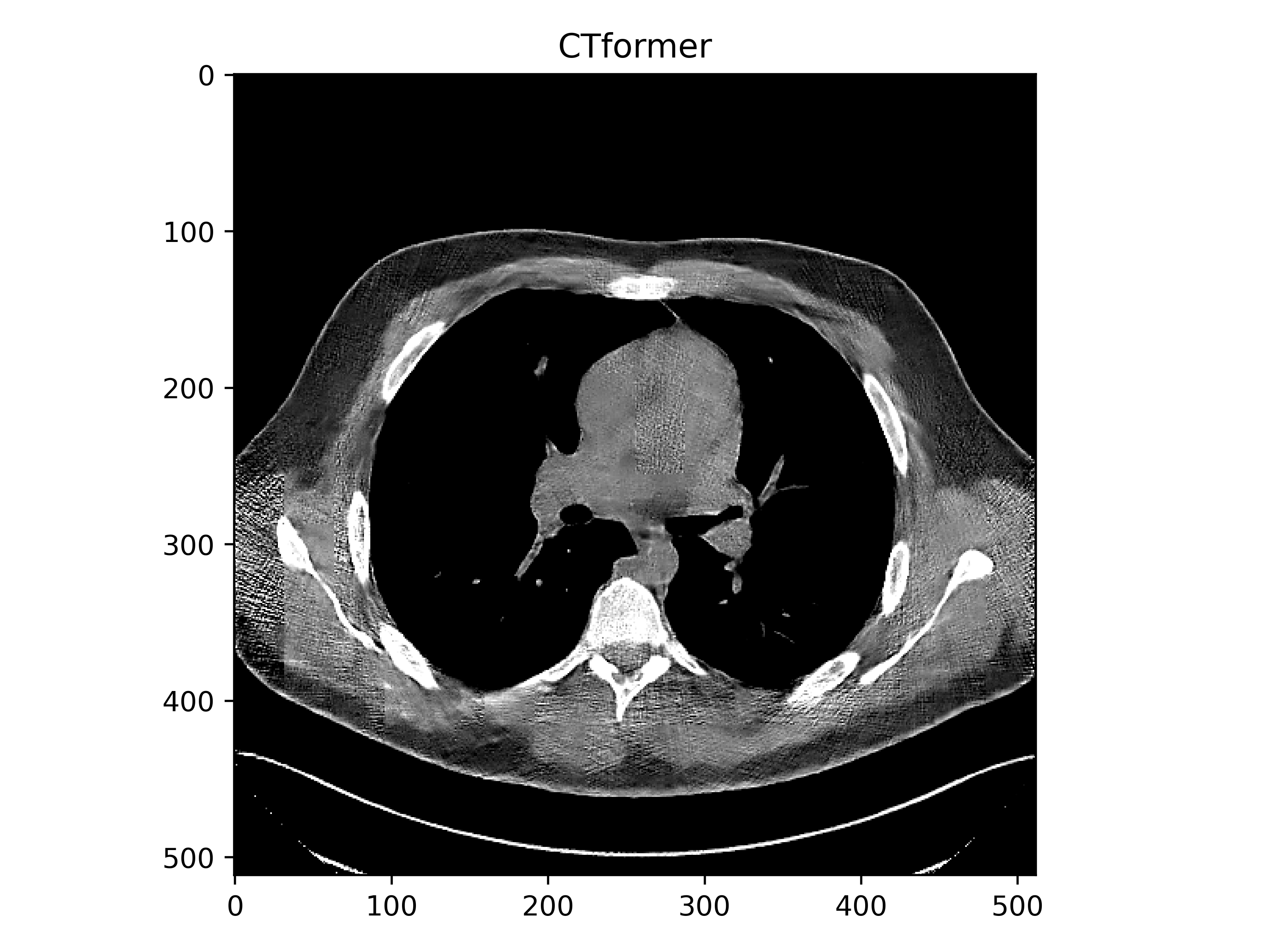}
	}
	\subfigure[]{
		\includegraphics[width=0.22\textwidth,trim=50 10 80 10, clip]{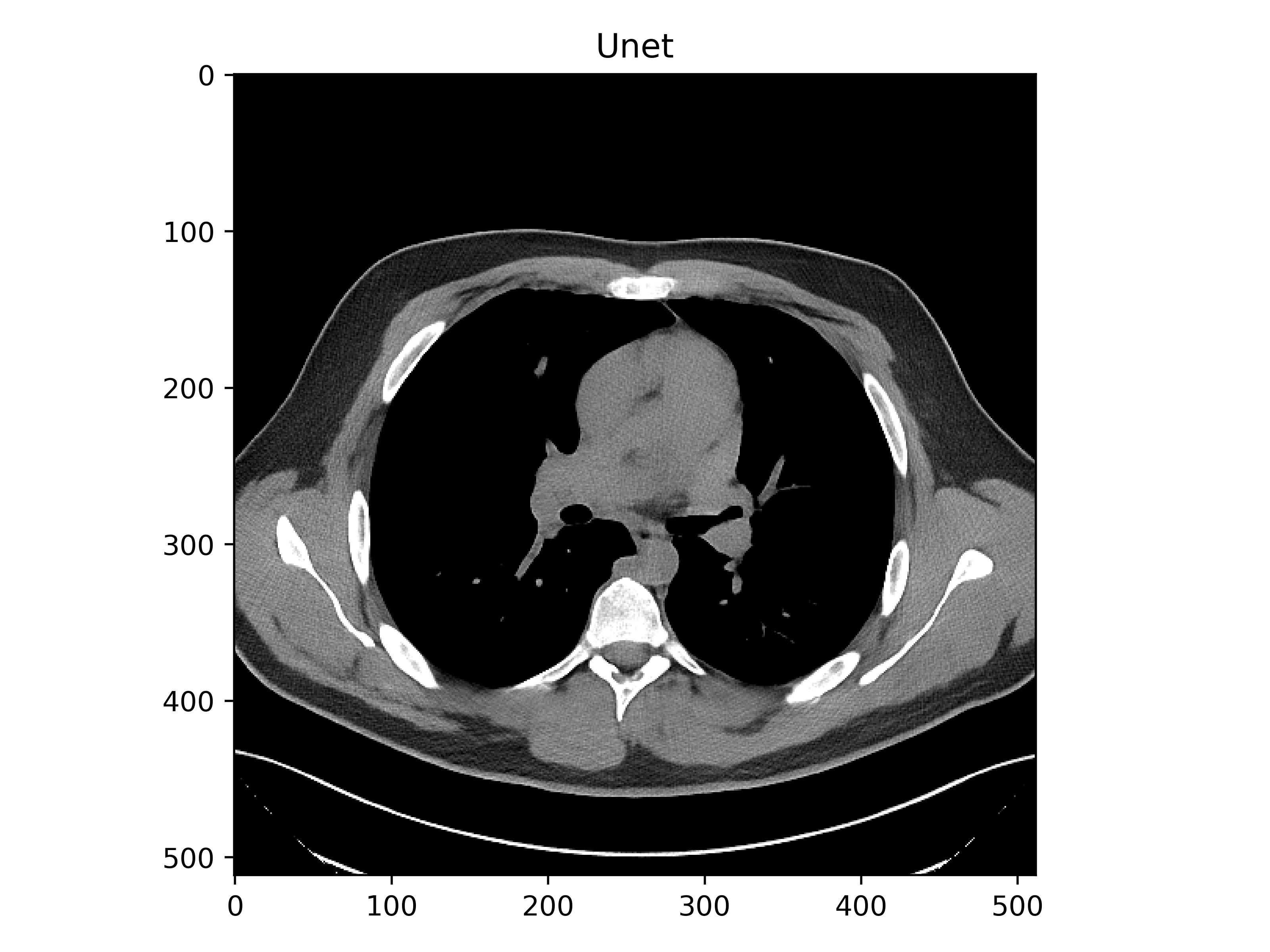}
	}
	\subfigure[]{
		\includegraphics[width=0.22\textwidth,trim=50 10 80 10, clip]{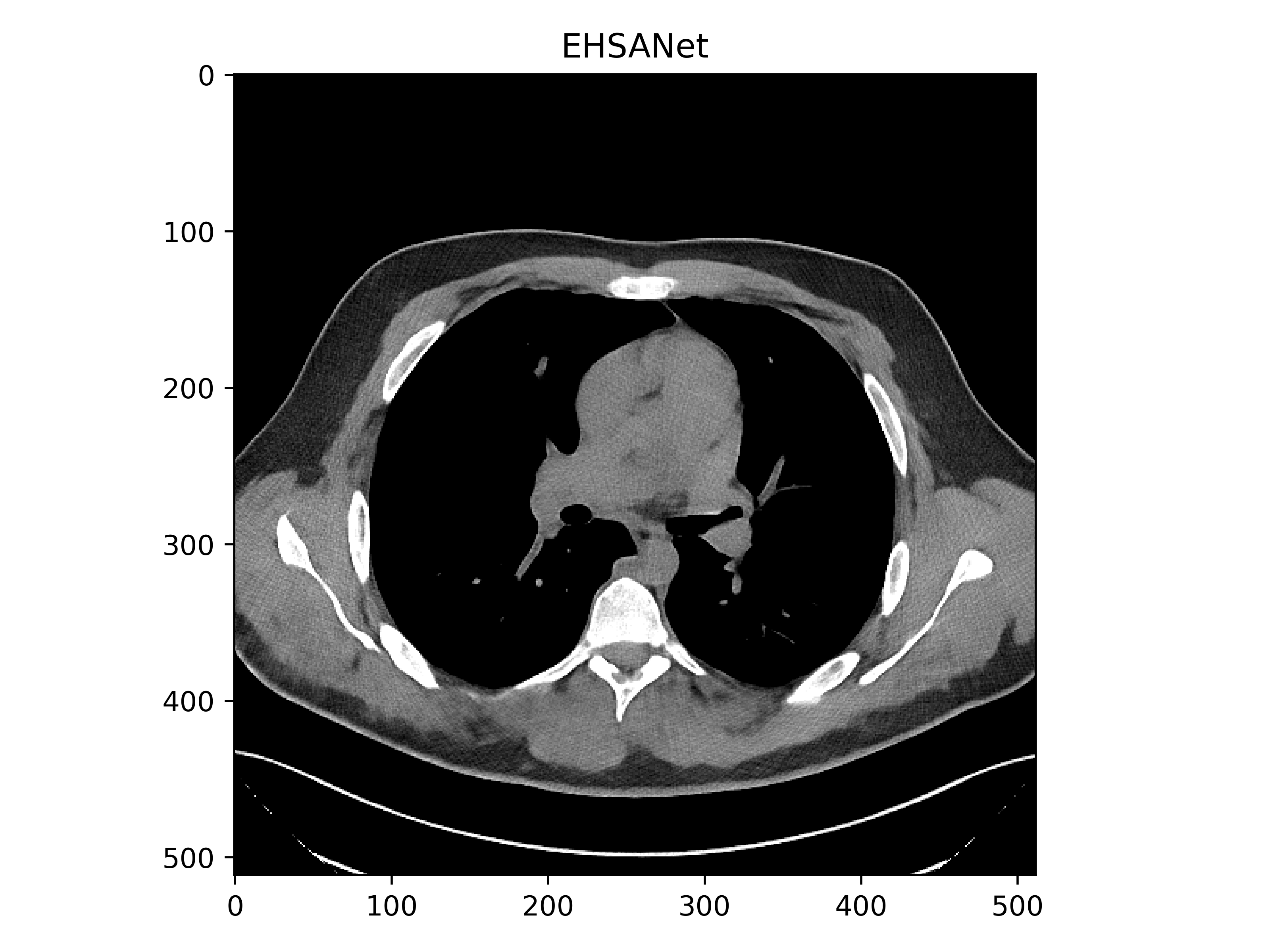}
	}
	\subfigure[]{
		\includegraphics[width=0.22\textwidth,trim=50 10 80 10, clip]{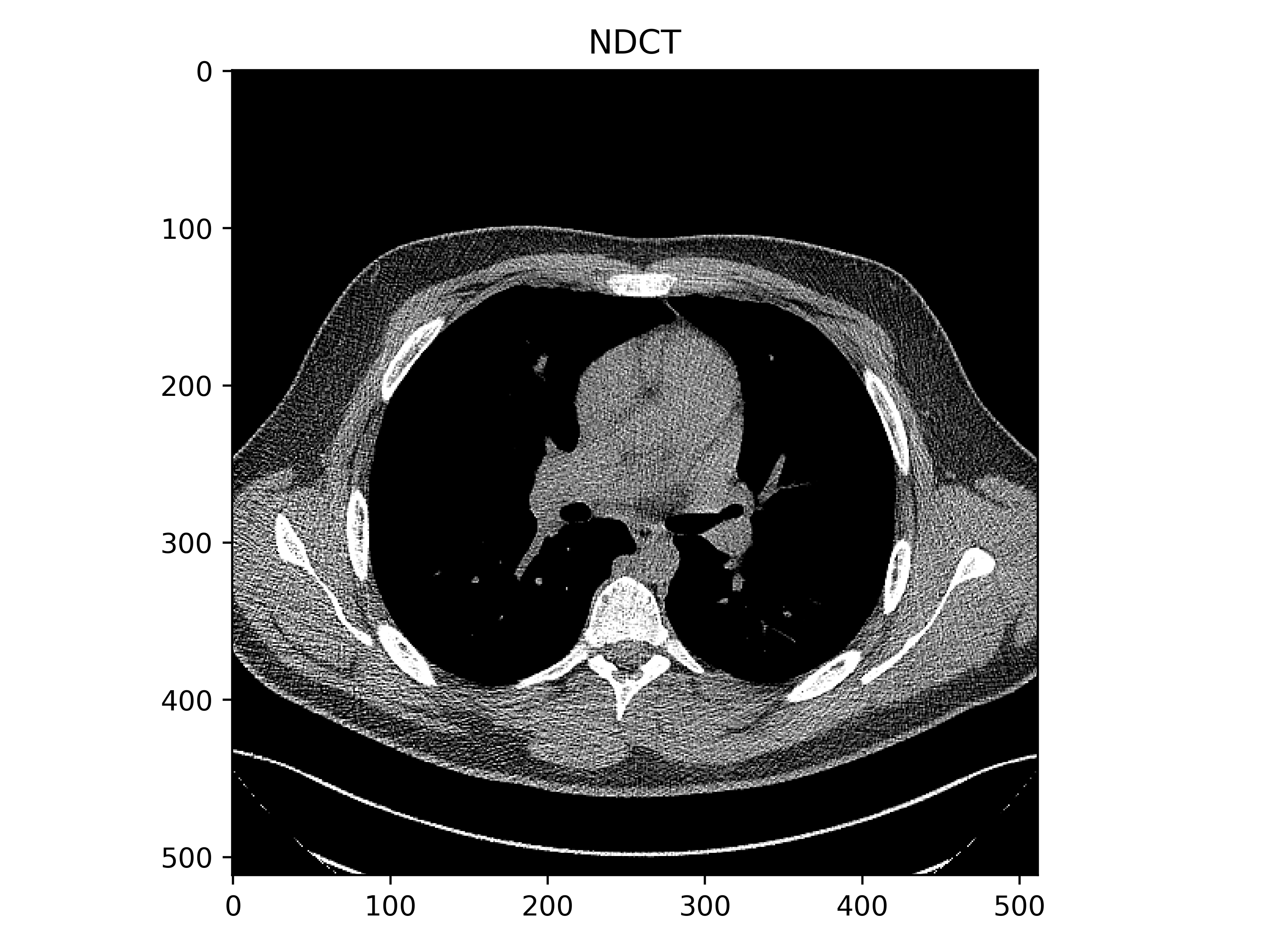}
	}
	
	\caption{Results of chest image for comparison. (a)LDPET, (b)RED-CNN,(c)Swin-Unet, (d)SwinIR, (e)CTformer, (f)Unet, (g)EHSANet large, (h)FDPET}
	\label{LDCT_chest_fig}
\end{figure*}

The results on the Mayo chest dataset are presented in Table \ref{comparison_ct_chest}. As shown, conventional CNN-based models, such as RED-CNN and U-Net, consistently achieve strong performance. An interesting observation is that the pure Transformer-based model, CTformer, attains performance comparable to RED-CNN on the LDCT abdomen dataset; however, on the Mayo chest dataset, it exhibits the worst performance among all evaluated methods. This discrepancy can be attributed to differences in anatomical structures and image characteristics between the two regions. Abdominal CT images are dominated by soft tissues, which generally exhibit low-frequency and relatively homogeneous intensity distributions. In contrast, chest CT images contain less soft tissue and are characterized by high-contrast regions, abundant high-frequency structures (e.g., lung textures and vascular details), and sparse intensity distributions. Such properties make noise and structural details more difficult to disentangle. Previous work by Kyung \textit{et al.} \cite{kyung2024generative} has shown that CTformer is prone to generating boundary artifacts, particularly in regions with sharp intensity transitions. This phenomenon is also observed in our experiments (Fig. \ref{LDCT_chest_fig}), where noticeable artifacts appear along structural boundaries in chest images. In comparison, our method achieves performance comparable to U-Net while maintaining significantly lower computational cost, as reflected by reduced FLOPs and parameter counts. This efficiency advantage is further illustrated in the patient-level distribution shown in Fig. \ref{LDCT_chest_violin}, where our approach demonstrates stable performance across cases.

Fig. \ref{LDCT_chest_fig} presents the denoised chest images. As observed, Swin-Unet and SwinIR introduce noticeable artifacts across the entire image. A similar issue is also evident in CTformer, particularly around structural boundaries, where the artifacts appear even more pronounced. In contrast, our method achieves performance comparable to U-Net, while producing cleaner results with reduced noise. Notably, the reconstructed images exhibit lower noise levels than the corresponding FDCT images.

\begin{table}[!h]
\centering
\begin{tabular}{lccccc}
\cline{1-6}
Methods & PSNR &SSIM & RMSE & GPU memory & parameters \\
\cline{1-6}
LDPET & 12.273 & 0.579 & 100.705 & - & - \\
RED-CNN & 19.792 & 0.670 & 43.079 & 5.08G & 1.85M \\
Swin-Unet & 19.269 & 0.651 & 45.497 & \textbf{1.2G} & 0.95M \\
SwinIR & 19.357 & 0.657 & 45.05 & 23G & \textbf{0.41M} \\
CTformer & 18.704 & 0.654 & 48.788 & 4.7G & 1.45M \\
Unet & \textbf{20.097} & \textbf{0.679} & \textbf{41.723} & 6G & 31M \\
EHSANet(large) & 20.062 & 0.678 & 41.903 & 10.4G & 3.88M \\
\cline{1-6}
\end{tabular}
\caption{Quantitative evaluation for LDCT chest dataset.}
\label{comparison_ct_chest}
\end{table}

\subsection{Scalability of Swin-Unet}
In Swin-Unet, the authors showed that increasing Swin-Transformer block numbers (network depth) does not significantly improve segmentation performance but considerably increases computational cost \cite{cao2022swin}. In line with Xiao \textit{et al.} \cite{xiao2021early}, who reported that the limited locality bias of Swin-Transformers makes them suboptimal for early feature extraction, we scaled up only the bottleneck stage using residual Swin-Transformer blocks while keeping the shallow Swin-Transformer layers unchanged, which led to improved performance. Furthermore, the number of convolutional blocks was increased from two to four in both the encoder and decoder, with skip connections added to improve feature reuse. The comparison detail shows in Table \ref{scalability}. Note that decay factor $\gamma$ here is set to 0.85 for scaled EHSANet model.

\begin{table}[!ht]
	\caption{scalability of EHSANet with residual Swin-Transformer bottleneck  }\label{scalability}
	\centering
	\begin{tabular}{lcc}
		\cline{1-3}
		Swin-Transformer bottleneck & \ding{51} &\ding{55}\\ 
		\cline{1-3}
		PSNR & 33.11 &32.88\\
		\cline{1-3}
	\end{tabular}
	
\end{table}

\section{Discussion and conclusion}
In this study, we developed EHSANet, a novel hybrid architecture designed to address the persistent challenge of the trade-off between training stability and computational efficiency in low-dose PET and CT imaging. By integrating EGA modules with a HIC upsampling strategy, our framework successfully balances noise suppression with the preservation of critical anatomical textures.

Our experimental results provide clear evidence that EHSANet achieves state-of-the-art performance, delivering the highest PSNR and lowest RMSE among compared methods while demonstrating remarkable stability across multiple training runs, especially on LDCT dataset. Beyond accuracy, the model’s lightweight design and efficient GPU memory usage make it a highly practical solution for real world deployment, where limited computational resources often hinder the use of large-scale Transformers

While HSANet effectively captures local and long-range dependencies, its scalability across deeper architectures has not yet been fully evaluated due to current computational constraints. Future work should investigate the impact of stacking additional hierarchical bottleneck structures to further enhance feature representation. In addition, we aim to develop noise-aware attention architectures that explicitly model the Poisson nature of PET data. Such advancements will be essential for further improving the reliability of image reconstruction in low-uptake regions, ensuring even greater diagnostic fidelity in ultra-low-dose protocols.

\section*{Acknowledgments}
Data used in the preparation of this article were obtained from the University of Bern, Department of Nuclear Medicine, and the School of Medicine, Ruijin Hospital. The investigators at these institutions contributed to the design and implementation of the data collection and/or provided the data but did not participate in the analysis or writing of this manuscript.

\section*{Data availibility}
The datasets are public available online. All deep learning methods were implemented using Pytorch (\url{https://pytorch.org/}). The custom script for this study will be available at \url{https://github.com/Christian-lyc/HSANet}.
\appendix




\bibliographystyle{elsarticle-num}
\bibliography{ref}



\end{document}